\documentclass[10pt,twocolumn,letterpaper]{article}

\usepackage[pagenumbers]{cvpr} %

\usepackage[dvipsnames, table, xcdraw]{xcolor}

\usepackage{graphicx}
\usepackage{siunitx}
\usepackage{booktabs}
\usepackage{tikz}
\usepackage{tabularx}
\definecolor{best_color}{RGB}{253,155,154}
\definecolor{second_color}{RGB}{254,205,158}
\definecolor{third_color}{RGB}{255,255,163} 

\definecolor{cvprblue}{rgb}{0.21,0.49,0.74}
\usepackage[pagebackref,breaklinks,colorlinks,citecolor=cvprblue]{hyperref}

\def\paperID{138} %
\def\confName{3DV\xspace}
\def\confYear{2026\xspace}

\title{CropCraft: Complete Structural Characterization of Crop Plants From Images}

\author{Albert J. Zhai$^{1}$ \and
Xinlei Wang$^{1}$ \and
Kaiyuan Li$^{1}$ \and
Zhao Jiang$^{1}$ \and
Junxiong Zhou$^{2}$ \and
Sheng Wang$^{1}$ \quad ~~~
Zhenong Jin$^{2}$ \quad ~~~
Kaiyu Guan$^{1}$ \quad ~~~
Shenlong Wang$^{1}$
\vspace{0.3em} \\
$^1$ University of Illinois Urbana-Champaign \quad
$^2$ University of Minnesota Twin Cities
}

\begin{document}

\maketitle
\begin{abstract}

The ability to automatically build 3D digital twins of plants from images has countless applications in agriculture, environmental science, robotics, and other fields. However, current 3D reconstruction methods fail to recover complete shapes of plants due to heavy occlusion and complex geometries. In this work, we present a novel method for 3D modeling of agricultural crops based on optimizing a parametric model of plant morphology via inverse procedural modeling. Our method first estimates depth maps by fitting a neural radiance field and then optimizes a specialized loss to estimate morphological parameters that result in consistent depth renderings. The resulting 3D model is complete and biologically plausible. We validate our method on a dataset of real images of agricultural fields, and demonstrate that the reconstructed canopies can be used for a variety of monitoring and simulation applications. Project page: {\url{https://ajzhai.github.io/CropCraft}}
\end{abstract}    
\section{Introduction}
\label{sec:intro}

Plants are ubiquitous objects that appear all around the world and serve as the foundation for agriculture, which underpins our civilization's growth and survival. The ability to automatically build 3D digital twins of plants from images has countless applications in agriculture, environmental science, robotics, and other fields. In particular, the development of such methods in the context of agriculture will enable automatic, large-scale monitoring of crops. 
The collected data can provide decision support for farmers, aid carbon budgeting for decision-makers, support the development of new agricultural techniques, and inform the design of new genotypes~\cite{zhu2020analysing, friedlingstein2023global, slattery2021perspectives}. All of these advances will contribute to increasing crop productivity, alleviating the rising food crisis of today's world~\cite{chand2008global, world2022state}.

\begin{figure}[!t]
\centering
\includegraphics[width=\linewidth,]{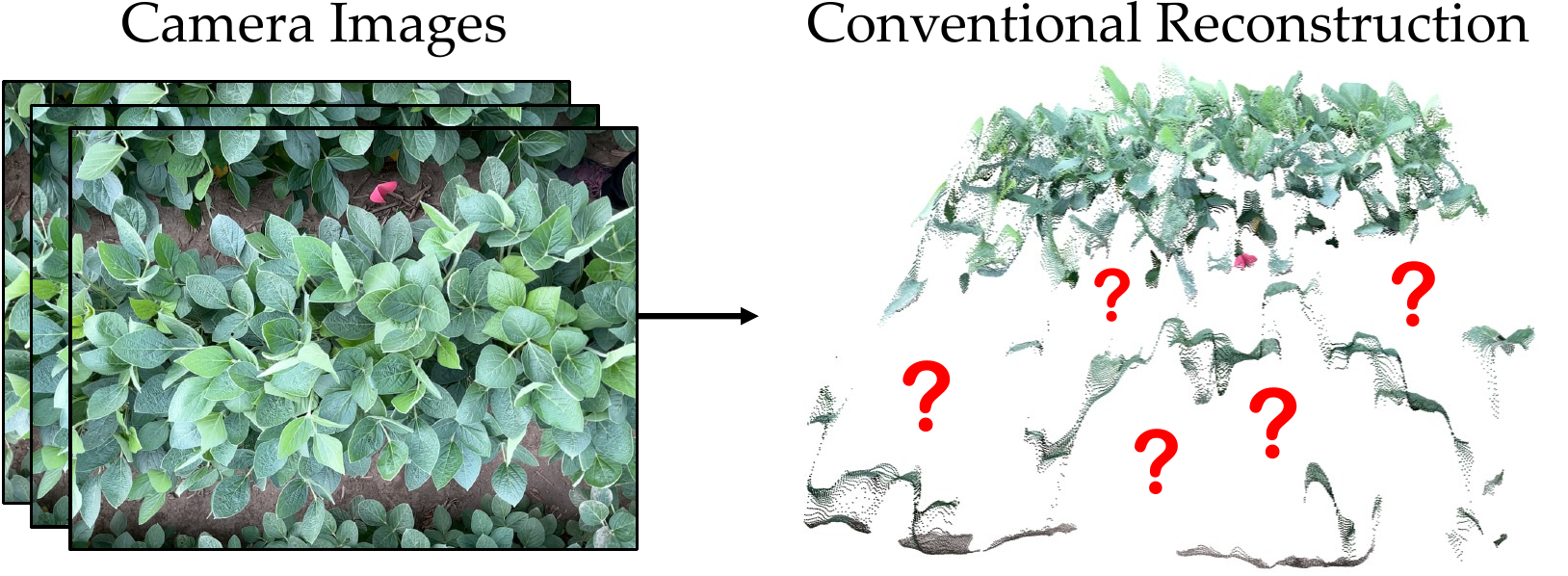}
\vspace{-14pt}
\caption{\textbf{Occlusion makes complete plant reconstruction challenging.} Current 3D reconstruction methods (e.g. VGGT~\cite{wang2025vggt}) miss a large portion of the crop canopy due to heavy occlusion, making them unsuitable for monitoring and analysis applications. We propose a novel method that combines neural rendering and procedural modeling to output a complete, interpretable, and biologically plausible 3D mesh model of the crop canopy.}
\vspace{-10pt}
\label{fig:teaser}
\end{figure}

However, 3D reconstruction of plants remains to be a challenging vision problem. Many plants, including most of those found in agriculture, are composed of complex arrangements of thin leaves and branches that heavily occlude one another. These low visibility conditions cause existing reconstruction pipelines to fall short. Multi-view reconstruction methods such as those based on neural rendering~\cite{mildenhall2020nerf, tancik2023nerfstudio, wang2021neus, yariv2021volume, li2023neuralangelo} or multi-view stereo~\cite{yao2020blendedmvs, seitz2006comparison, wei2021aa, zhang2023vis} do not reconstruct the invisible regions of the scene, leading to incomplete plant shapes and extraneous geometry. Learning-based methods~\cite{hong2023lrm, liu2023zero, gao2024cat3d} can overcome this issue but require large amounts of ground-truth 3D data for training, which is extremely difficult to collect for dense vegetation due to the same occlusion reasons.

On the other hand, there is a large body of work that has found success in modeling 3D plant shapes via procedural generation~\cite{li2021learning, prusinkiewicz1999system, longay2012treesketch, stava2014inverse, song2020decomposition, qian2023coupled}. These procedural models are grounded in scientific knowledge and are carefully designed to produce biologically plausible plant shapes that consist of anatomically complete arrangements of plant organs with realistic shapes. However, these models generally require human input to set their parameters, and the task of automatically generating plants that accurately represent plant instances observed in the real world (``inverse procedural modeling'') remains to be difficult.

In this work, we present a novel method for 3D modeling of agricultural crops based on optimizing the parameters of a procedural plant morphology model. Our method combines the flexibility of data-driven neural reconstruction methods with the robust foundational knowledge in procedural models. To this end, we first use neural radiance field (NeRF) techniques~\cite{mildenhall2020nerf, tancik2023nerfstudio} to estimate the geometry of the visible surfaces in the scene. We then apply heuristics to localize the planting rows and calculate a \textit{row-aligned camera pose} from which to render depth maps. Next, we render depth maps from both the NeRF and a virtual scene generated by our procedural model, and use Bayesian optimization to minimize a specially designed loss function with respect to the procedural model's parameters. 

The design of the loss function and the parameterization of the procedural model are crucial for obtaining a 3D canopy model that is useful for field-level analysis applications. Importantly, for such applications, the fine-grained positions of each \textit{individual} leaf are mostly irrelevant, as long as the \textit{aggregate shape characteristics} of the crop canopy are faithful. We thus define our loss on \textit{histogram statistics} of the depth map and optimize a highly \textit{compact} set of parameters to ensure that the key canopy characteristics for determining crop productivity are accurately captured without being distracted by fitting irrelevant details. Thanks to the strong priors imposed by the procedural model, this process jointly optimizes the \textit{complete} plant shape, including portions not visible in the input images.

To validate our approach, we collect a multi-view dataset of real soybean and maize fields, paired with manual measurements of leaf area and leaf angle. We show that the proposed method can successfully reconstruct realistic crop canopies across different growth stages and estimate key canopy structure variables more accurately than baselines. These variables are tightly linked to crop yield and are used extensively in crop science. We also show that the reconstructed 3D canopies can be directly fed into radiative transfer modeling software to provide accurate predictions of photosynthesis rates. The results highlight the potential for monitoring crop productivity directly from camera images instead of costly flux tower equipment. Our contributions are summarized below: 
\begin{itemize} 
\item We present a novel approach for reconstructing complete 3D morphological models of large-scale crop plant fields from a collection of images. 
\item We provide a framework for image-based growth and photosynthesis quantification, paving the way for scalable yield prediction and carbon exchange monitoring.
\end{itemize}

\section{Related Work}
\label{sec:related}

\paragraph{3D Reconstruction.}
The task of reconstructing the 3D geometry of a scene given images is a longstanding problem in computer vision. Most reconstruction pipelines start from Structure-from-Motion (SfM), which infers camera parameters for the input images through joint optimization with detected keypoints~\cite{triggs2000bundle, schonberger2016structure}. Afterwards, 3D geometry can be estimated based on photometric consistency across input views. Multi-view stereo (MVS) methods attempt to match correspondences across images and then triangulate the 3D coordinates of the points~\cite{yao2020blendedmvs, seitz2006comparison, wei2021aa, zhang2023vis}. Another approach for 3D reconstruction is based on fitting a neural radiance field (NeRF) to the scene~\cite{mildenhall2020nerf, tancik2023nerfstudio, wang2021neus, yariv2021volume, li2023neuralangelo, lin2021barf, yariv2023bakedsdf}. NeRF is a high-fidelity and compact 3D scene representation that consists of a color field and a density field, both parameterized by neural networks and fitted to the input images via differentiable volume rendering~\cite{mildenhall2020nerf}. After training, depth maps can be obtained by volume rendering with the density field. NeRF-based methods can model view-dependent effects and tend to give better surface geometry than MVS. More recently, new methods based on 3D Gaussians have been proposed~\cite{kerbl20233d, guedon2023sugar, huang20242d}, achieving higher visual fidelity and efficiency than NeRF, but often with less accurate geometry. In our approach, we use an off-the-shelf NeRF~\cite{tancik2023nerfstudio} to reconstruct visible surfaces of plants, which has been explored before~\cite{smitt2023pag, arshad2024evaluating, saeed2023peanutnerf, hu2023high}. However, due to the inevitable occlusion in agricultural scenes, this is inadequate for full canopy reconstruction. To solve this issue, we turn to procedural models, which incorporate scientific knowledge and define a space of complete plant shapes that we can constrain our solution to.

\begin{figure*}[!t]
\centering
\includegraphics[width=\linewidth, trim={0 0cm 0 0cm}, clip]{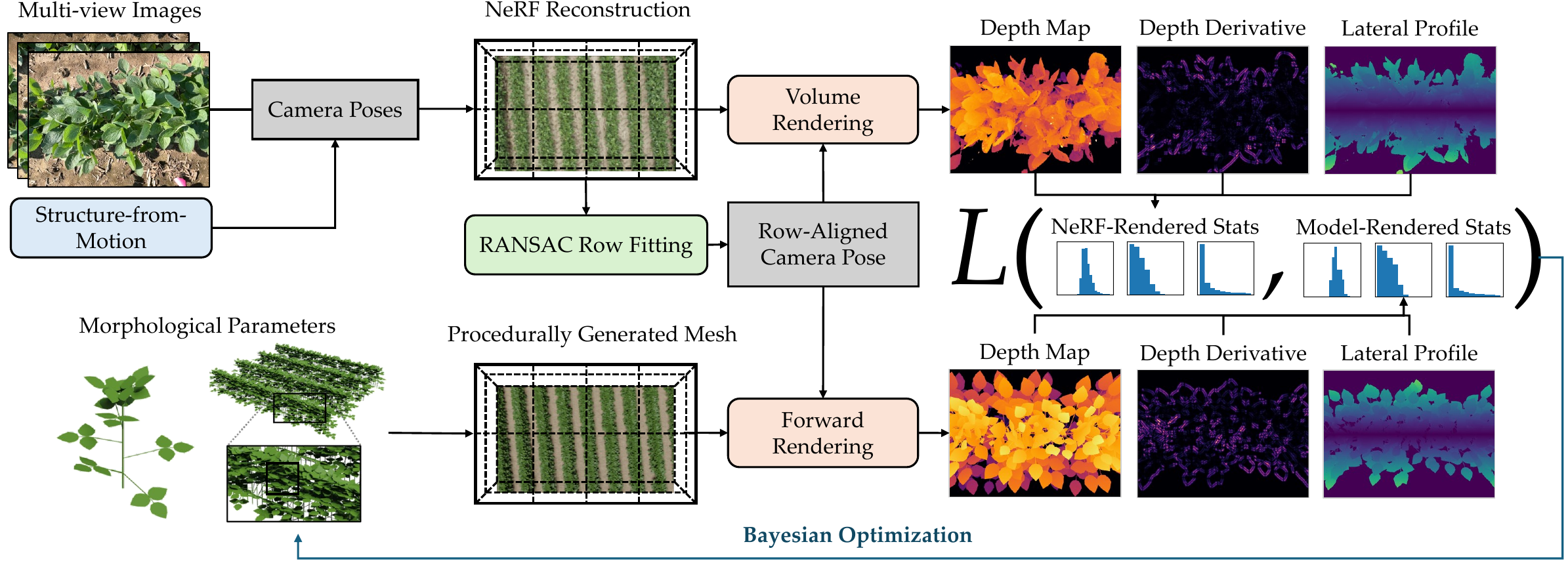}
\vspace{-12pt}
\caption{\textbf{Overview of our method.} We aim to estimate the parameters for a procedural generation model to generate a shape that matches the observed images. First, we use standard structure-from-motion and neural radiance field (NeRF) techniques to reconstruct the visible geometry of the scene. We then apply geometric heuristics to calculate a camera pose aligned with the planting rows of the crops. This pose is used to render depth maps from both the NeRF and the procedural model. We define a specialized loss function based on histogram statistics of the depth maps and minimize it with respect to the morphological parameters using Bayesian optimization.}
\label{fig:full_overview}
\end{figure*}   

\vspace{-8pt}

\paragraph{Inverse Procedural Modeling.}
Inverse procedural modeling (IPM) refers to the problem of finding a procedure for generating a 3D representation of a given object or scene. The key design choices for IPM pipelines are the choice of procedural model and the method for optimizing the parameters of the model. IPM has been applied for various input categories, with works focusing on building interiors~\cite{furukawa2009reconstructing, chen2019floor, ikehata2015structured, vid2cad, im2cad}, facades~\cite{martinovic2013bayesian, Fan2014StructureCF, wu2014inverse}, driving scenes~\cite{kar2019meta, devaranjan2020meta}, and forestry~\cite{stava2014inverse, guo2018realistic, niese2022procedural, li2021learning, li2024svdtree, liu2025neural}. The procedural models range from scene grammars~\cite{kar2019meta, ganeshan2023improving, im2cad} to constructive solid geometry trees~\cite{yu2024d}
to L-systems~\cite{prusinkiewicz1999system, palubicki2009self}, which are especially popular for modeling botanical trees. To fit the model, a structural similarity measure is defined based on geometric and or semantic agreement. Depending on the type of model, which may require estimating procedural rules and or parameters~\cite{aliaga2016inverse}, various search-based or dynamic programming-based optimization methods may be used to maximize the similarity measure. Unlike the reconstruction methods described above, inverse procedural modeling methods tend to focus less on photometrically exact reconstructions, aiming to obtain structurally similar approximations useful for simulation applications.

To the best of our knowledge, our work is the first to use IPM to reconstruct agricultural crops in the field and the first to integrate NeRFs into the IPM pipeline. Compared to individual plants~\cite{zhu2020analysing, liu2025neural}, modeling dense crop canopies poses a new set of challenges due to the high levels of occlusion. IPM methods for trees~\cite{stava2014inverse, li2024svdtree} are inspirational but do not adequately constrain the plant topology for smaller crops. We propose a specialized IPM approach to estimate 3D crop meshes that can be used for plant phenotyping, visualization, and simulation of biophysical processes.

\vspace{-8pt}

\section{Inverse Procedural Modeling for Crops}
\label{sec:method}

Our proposed method takes as input a set of images of a field of crops, and outputs a set of parameters that can be fed into a procedural generation model to produce a 3D mesh of the crops in the images. An overview of our method is provided in Fig.~\ref{fig:full_overview}. By constraining the space of possible 3D shapes to the output space of the procedural model, we can ensure that the reconstructed plants are complete and have realistic leaf shapes and branch topology. Of course, we also need our reconstruction to match what we can observe in the input images. We achieve this by searching for parameters that minimize a specialized loss function based on depth maps estimated from the input images and corresponding depth maps rendered from the reconstruction.

The layout for the rest of this section as follows. In Sec.~\ref{sec:procedural_model}, 
we describe the procedural generation models we use for plant morphology. In Sec.~\ref{sec:depth_rendering}, we describe how we use a neural radiance field (NeRF)~\cite{mildenhall2020nerf} and geometric heuristics to obtain row-aligned depth maps from the input images. In Sec.~\ref{sec:loss}, we describe our loss function and in Sec.~\ref{sec:opt}, we describe our optimization algorithm.

\subsection{Procedural Morphology Model}
\label{sec:procedural_model}
At the core of our method is a procedural morphology model that defines a primitive for the leaf shape and the possible topologies for the plant structure. Since this varies drastically across different species, these models should be species-dependent. In this work, we focus on soybean (\textit{Glycine max}) and maize (\textit{Zea mays}), the two most-grown crops in the United States~\cite{nass2021nass}. 

\begin{figure}[!t]
\centering
\includegraphics[width=\linewidth, trim={0 5.5cm 0 1cm}, clip]{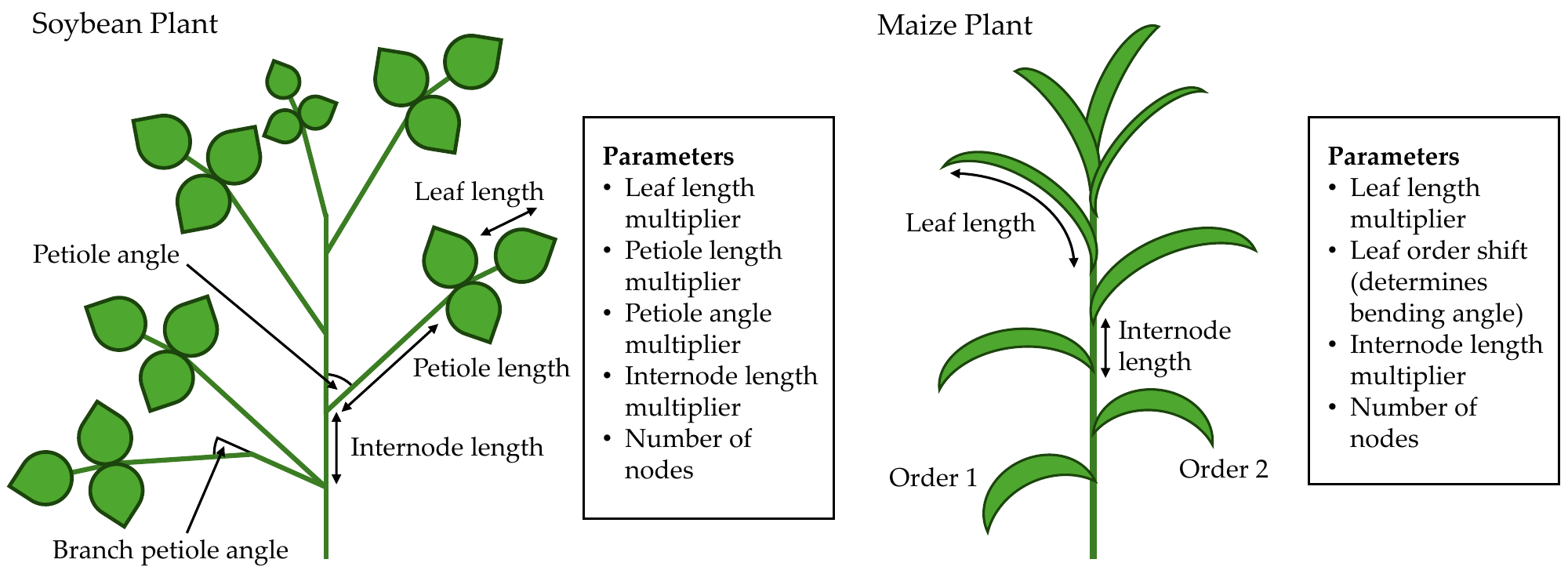}    
\vspace{-8pt}
\caption{\textbf{Procedural plant morphology models.}   We adopt procedural generation models from existing soybean morphology~\cite{song2020decomposition} and maize morphology~\cite{qian2023coupled} models. This (stylized) illustration shows the parameters that we allow to be optimized to model variations across individual instances of each species. }
\vspace{-8pt}
\label{fig:morphology}
\end{figure}

Accurate crop modeling requires appropriate species-based shape constraints. Our procedural generation models are heavily based on existing work in the crop morphology literature. For soybean, we build upon the mCanopy model by Song et al.~\cite{song2020decomposition}, and for maize, we build upon the coupled maize model by Qian et al.~\cite{qian2023coupled}. The mCanopy model consists of nodes of trifoliate leaves attached via a petiole to either the main plant stem or a branch stem. The maize model consists of large curved leaves growing from a single main stem. We introduce our own parameterization for each model, as illustrated in Fig.~\ref{fig:morphology}. The parameterization is designed to be flexible enough to capture the primary shape variations across instances of each species while remaining as low-dimensional as possible (5 parameters for soybean, 4 for maize). Note that the procedural generation process is in general stochastic; e.g. the model samples leaf angles according to a probability distribution. We refer readers to the supplementary for additional details about each model.

\subsection{Pose Alignment for Depth Rendering}
\label{sec:depth_rendering}
The procedural generation model defines a space of 3D shapes that can represent plants of various stages of maturity, growing conditions, and cultivars. In order to search within this space for the shape that best matches the plants in input observations, we define a specialized loss function based on depth maps estimated from the input images and depth maps rendered using the proposed shape. 

Note that, in order to properly compare depth maps between the real scene and the procedurally generated scene, the procedural plants should be placed in the same positions and orientations (relative to the camera) as in the real scene. However, it is practically impossible to determine the precise base position of every individual plant under heavy occlusion. Fortunately, it is also \textit{unnecessary} to do so, since our applications demand characterizing the canopy overall, not each individual plant. Thus, we handle pose alignment by using geometric heuristics to estimate plant \textit{row} locations, while carefully designing our loss function to be invariant to individual plant locations \textit{within} each row.

We assume that our crop plants are planted in rows with a known planting density (typically true). To estimate the row locations, we first use an off-the-shelf NeRF (Nerfacto~\cite{tancik2023nerfstudio}) to obtain a 3D point cloud of the visible surfaces. We render the depth for each input view and unproject each pixel to a 3D point. Then, we randomly sample a fixed number of points within a bounding box for the scene and perform voxel downsampling. Next, we segment the points into ground/plant using a threshold on their color in LAB space. We fit a plane to the ground points using RANSAC~\cite{schnabel2007efficient} and take an upper slice of the plant points based on their distance to the ground plane. Finally, we fit a line to each row by sequentially running RANSAC and removing the inliers for each fitted line. Additional details for the row-fitting procedure are provided in the supplementary.

Given the ground plane and rows, we define a standardized camera to render from at a predetermined height above the center of the row with the most inliers, facing straight downwards with its $x$-axis aligned with the row. This design will allow our loss to be invariant to the exact placement of plants along the $x$-axis. The camera is used to render both the observation depth map $\mathbf{I}_{\text{obs}}$ from the NeRF and the depth map $\mathbf{I}_{\text{pred}}$ induced by the procedurally generated scene.  Note that depth estimators such as VGGT~\cite{wang2025vggt} are unsuitable here since we are rendering depth from a novel (non-input) view. We also use a color threshold and a bounding box to acquire foreground (plant) masks $\mathbf{M}_{\text{obs}}$ and $\mathbf{M}_{\text{pred}}$ corresponding to each depth map.

\begin{table*}[!t]
    \centering
    \resizebox{\linewidth}{!}{
\setlength{\tabcolsep}{0.2em} %
\renewcommand{\arraystretch}{1.}
    \begin{tabular}{cccc}
    Input & Observation Depth & Predicted Depth & 3D Mesh \\
          \tikz{
        \node[draw=black, line width=.5mm, inner sep=0pt] 
            {\includegraphics[width=.25\linewidth]{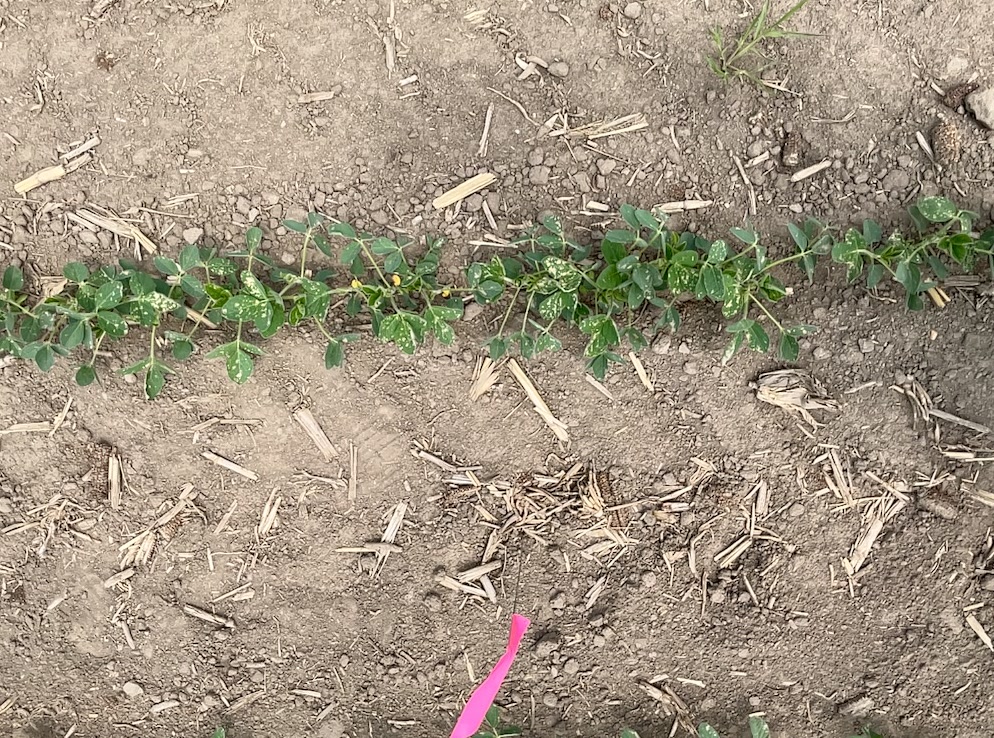}};
            \node[draw=black, draw opacity=1.0, line width=.3mm, fill opacity=0.8,fill=white, text opacity=1] at (-1.4 , 1.25) { June 16 };
        } &
        \tikz{
        \node[draw=black, line width=.5mm, inner sep=0pt] 
            {\includegraphics[width=.25\linewidth]{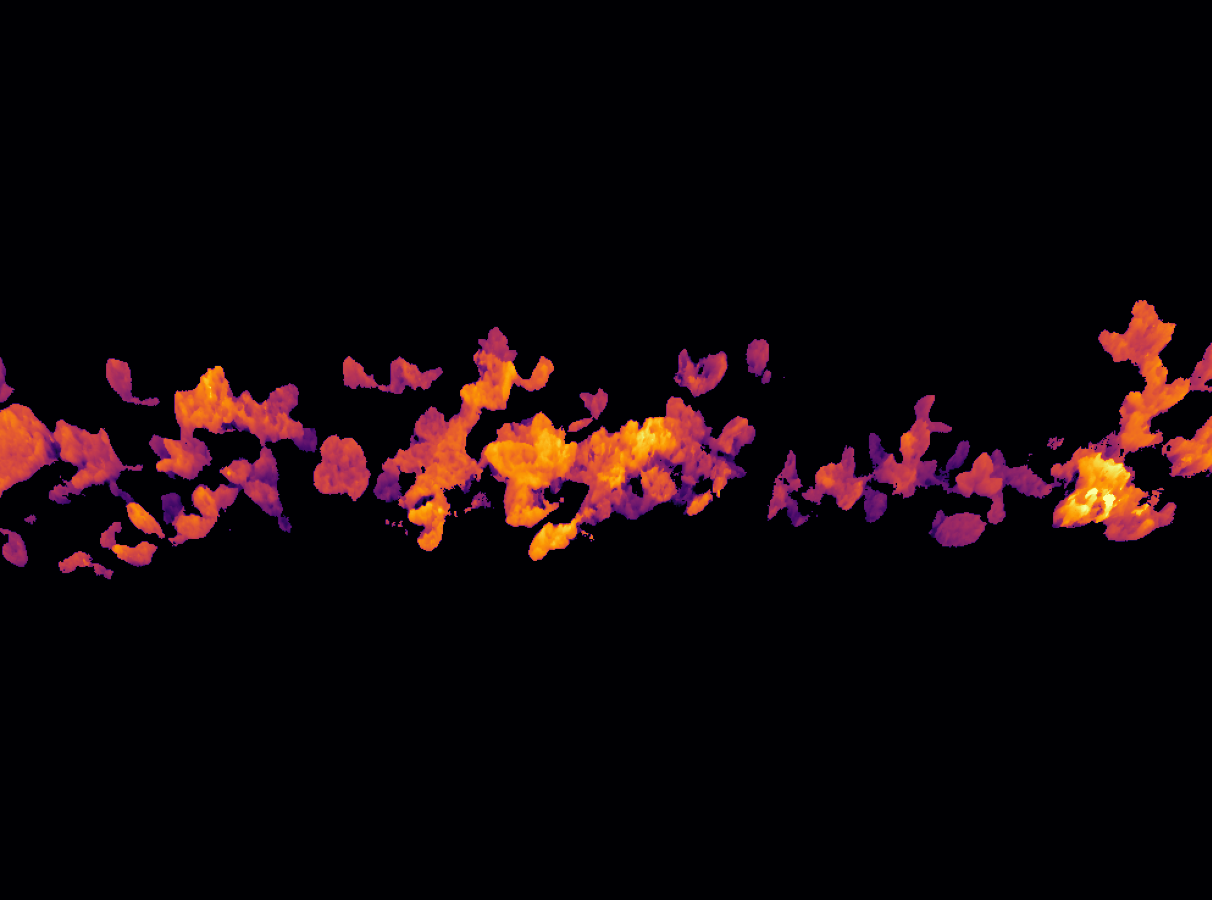}};
            ;
        }& 
        \tikz{
        \node[draw=black, line width=.5mm, inner sep=0pt] 
            {\includegraphics[width=.25\linewidth]{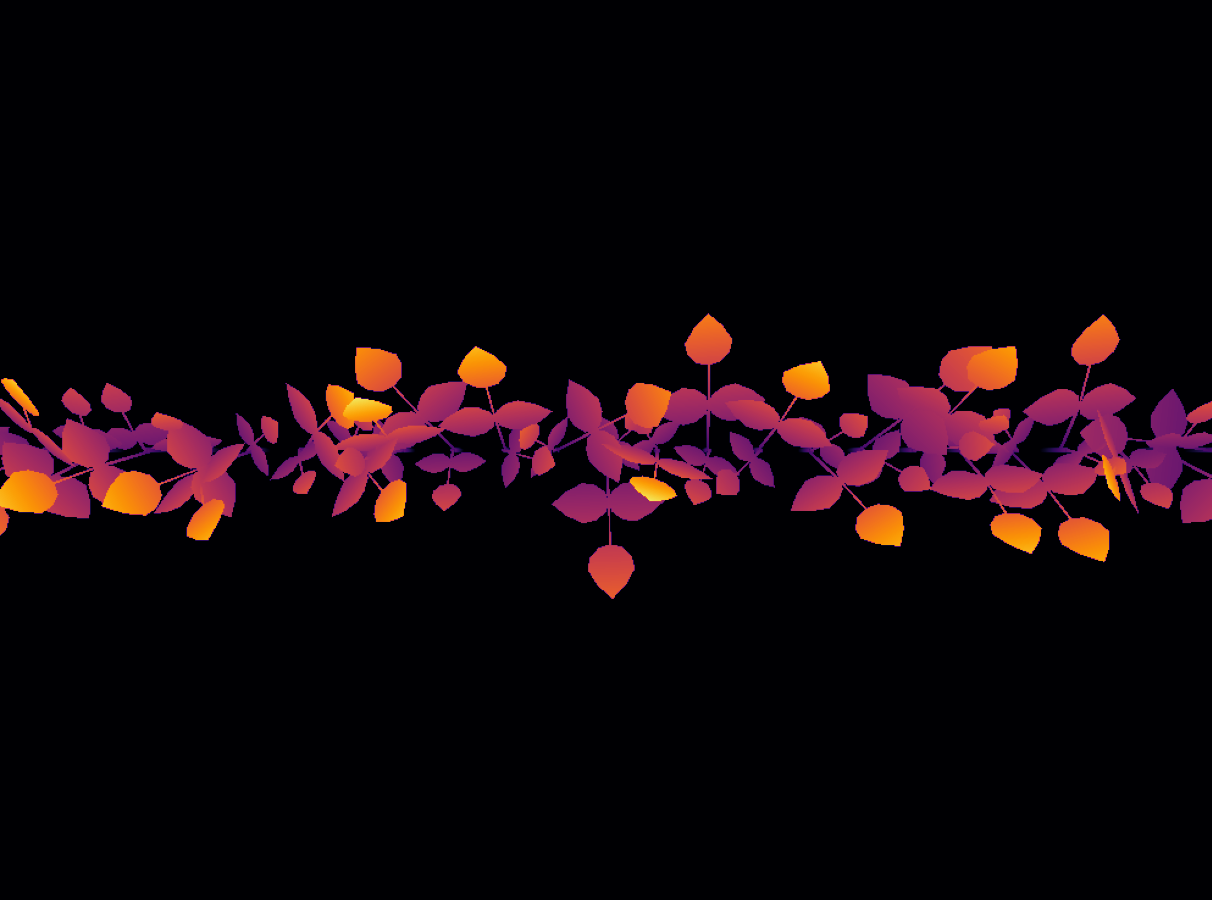}};
            ;
        }& 
       \tikz{
        \node[draw=black, line width=.5mm, inner sep=0pt] 
            {\includegraphics[width=.25\linewidth]{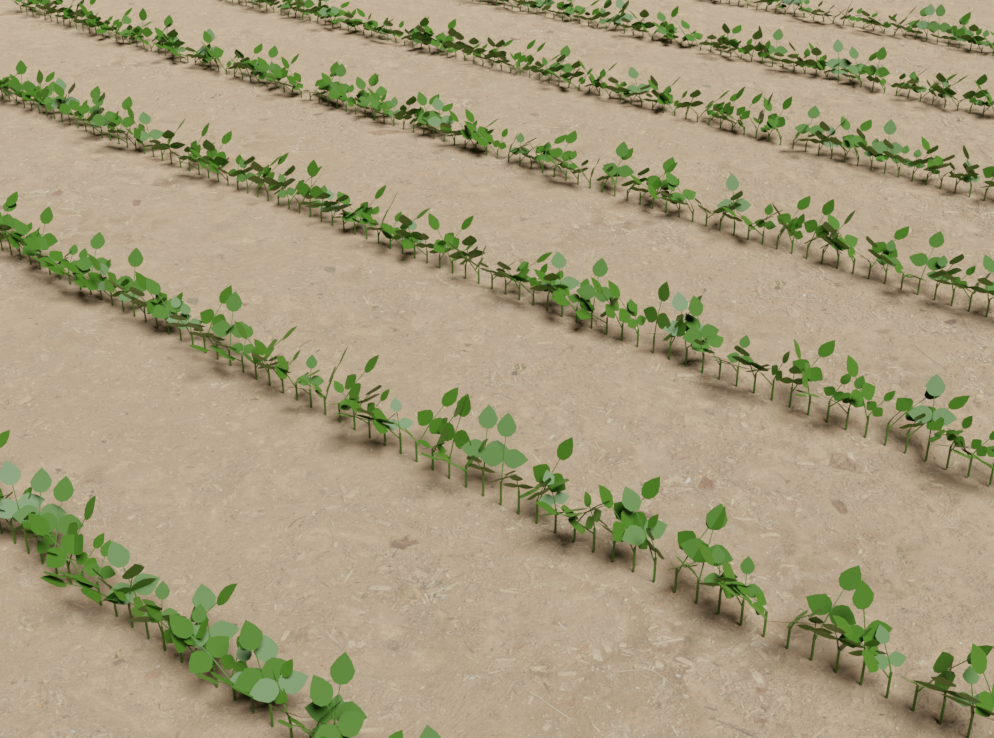}};
            ;
        } 
        \\
        \tikz{
        \node[draw=black, line width=.5mm, inner sep=0pt] 
            {\includegraphics[width=.25\linewidth]{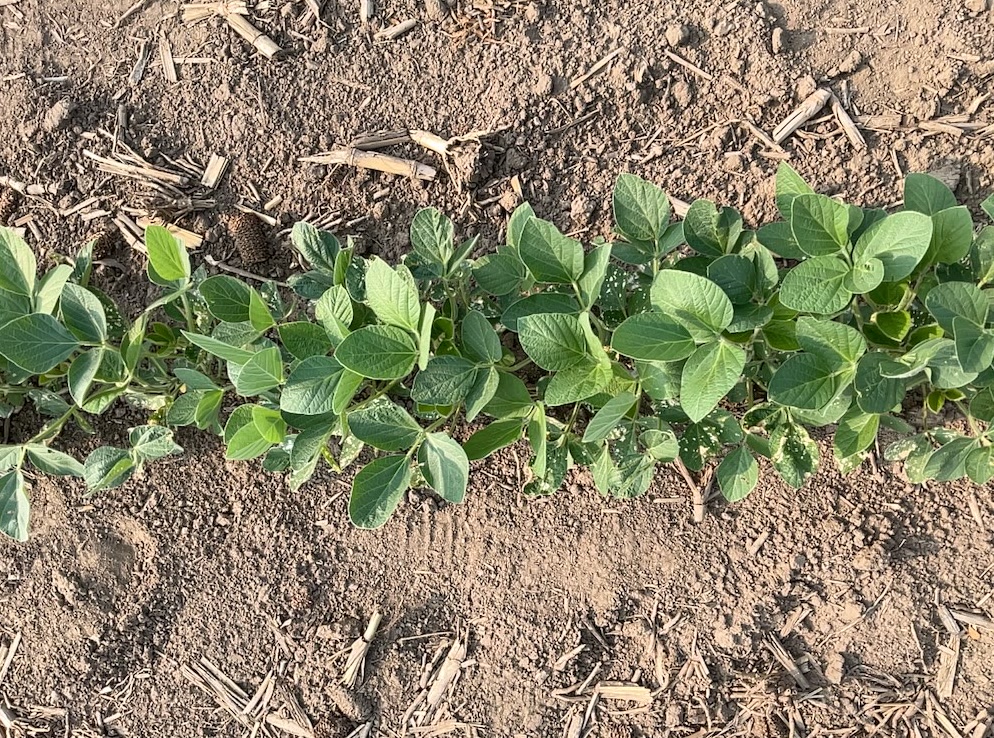}};
            \node[draw=black, draw opacity=1.0, line width=.3mm, fill opacity=0.8,fill=white, text opacity=1] at (-1.4 , 1.25) { June 27 };
        } &
        \tikz{
        \node[draw=black, line width=.5mm, inner sep=0pt] 
            {\includegraphics[width=.25\linewidth]{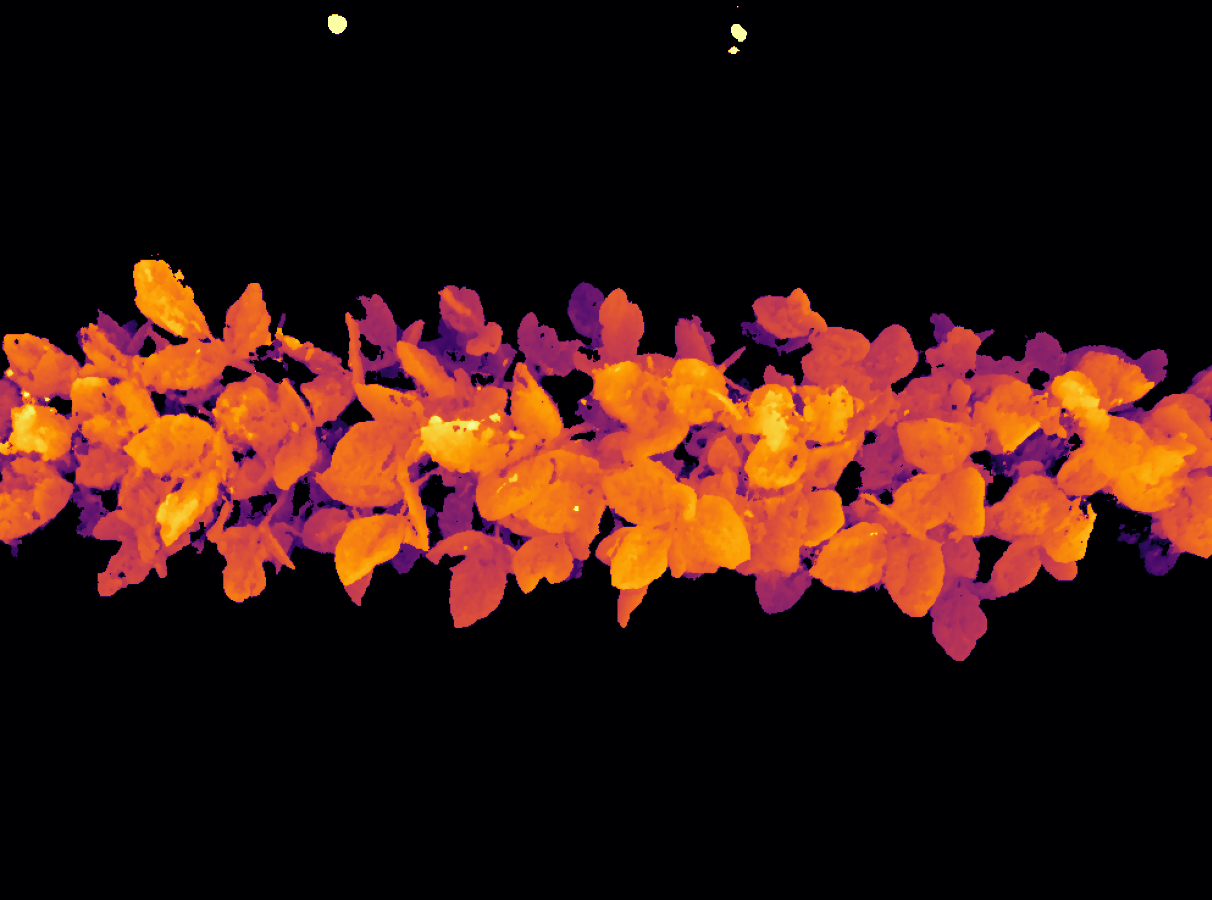}};
            ;
        }& 
        \tikz{
        \node[draw=black, line width=.5mm, inner sep=0pt] 
            {\includegraphics[width=.25\linewidth]{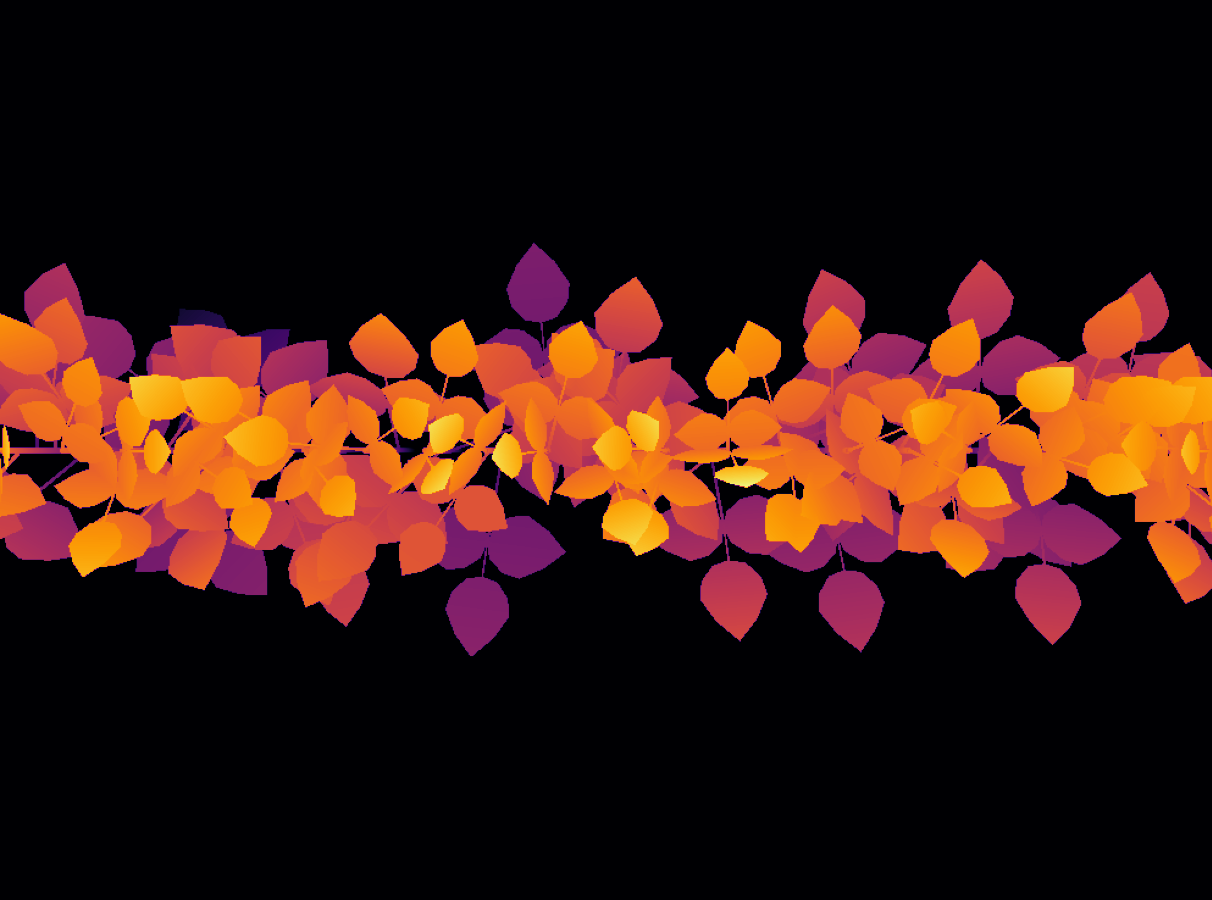}};
            ;
        }& 
       \tikz{
        \node[draw=black, line width=.5mm, inner sep=0pt] 
            {\includegraphics[width=.25\linewidth]{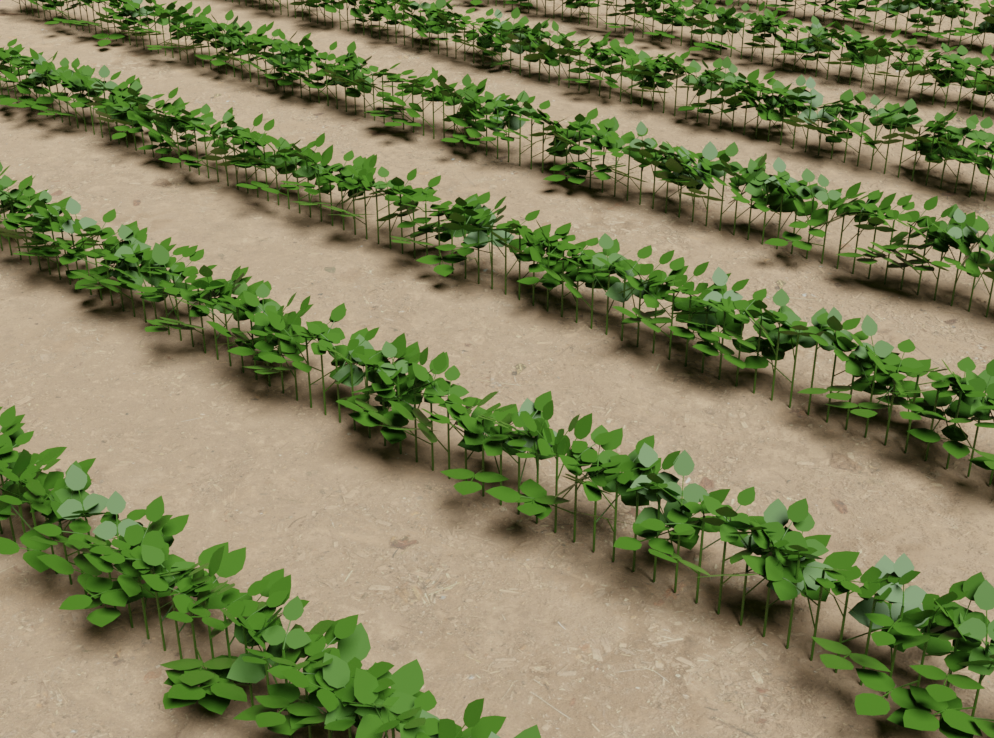}};
            ;
        } \\

        \tikz{
        \node[draw=black, line width=.5mm, inner sep=0pt] 
            {\includegraphics[width=.25\linewidth]{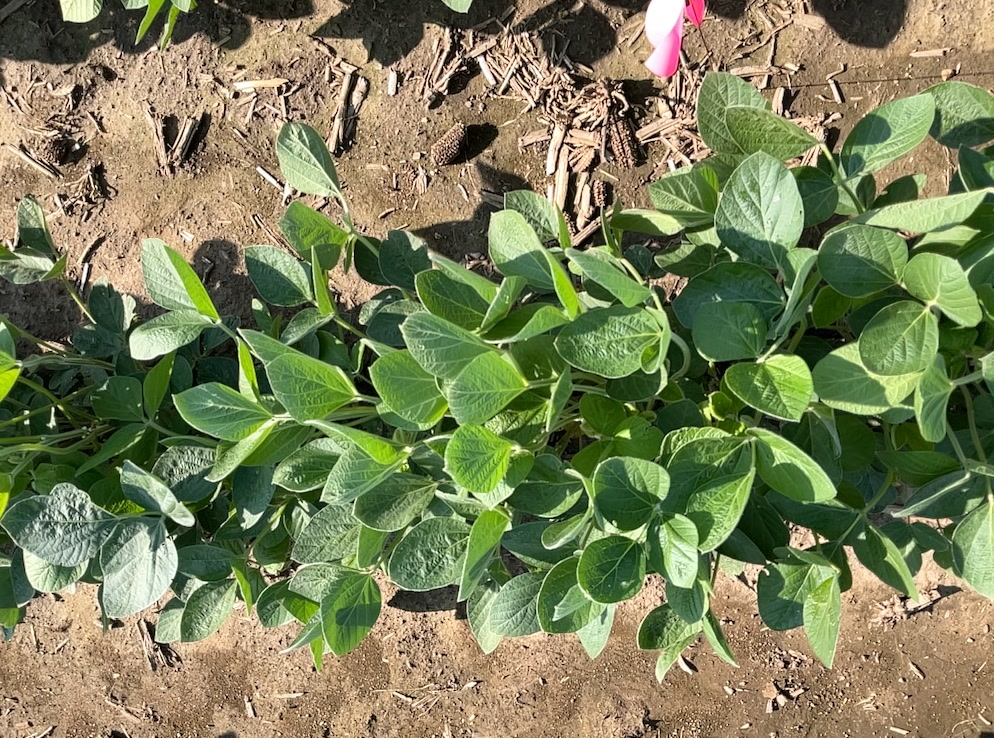}};
            \node[draw=black, draw opacity=1.0, line width=.3mm, fill opacity=0.8,fill=white, text opacity=1] at (-1.45 , 1.25) { July 11 };
        } &
        \tikz{
        \node[draw=black, line width=.5mm, inner sep=0pt] 
            {\includegraphics[width=.25\linewidth]{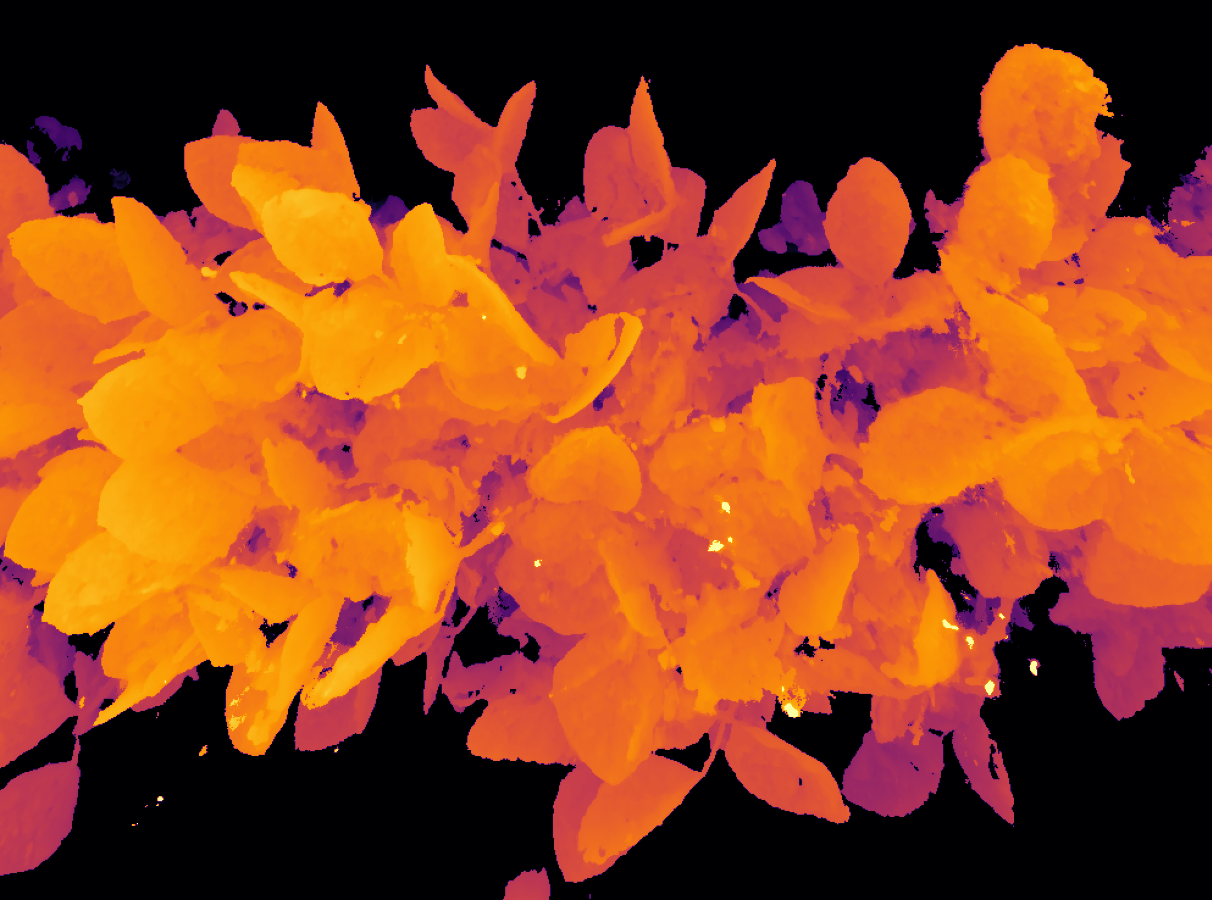}};
            ;
        }& 
        \tikz{
        \node[draw=black, line width=.5mm, inner sep=0pt] 
            {\includegraphics[width=.25\linewidth]{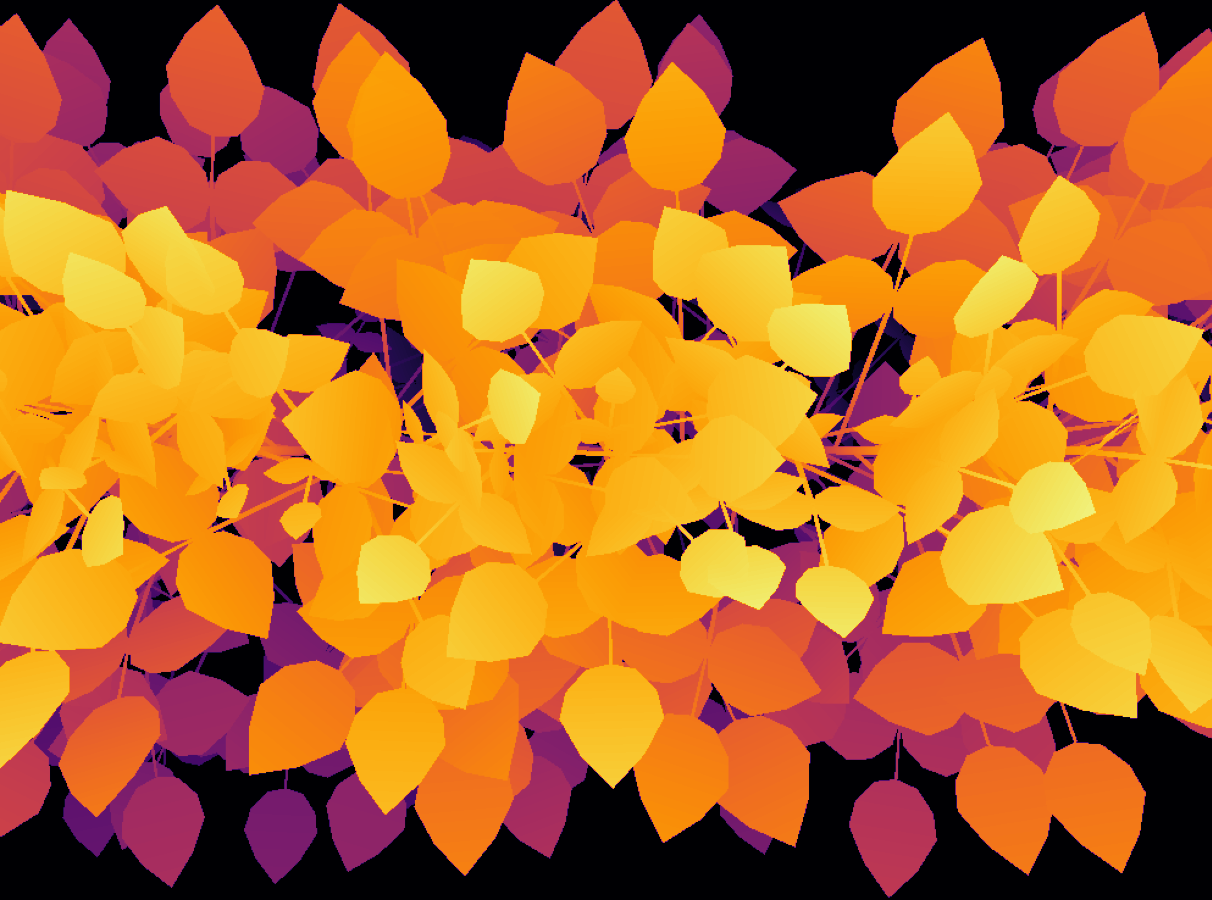}};
            ;
        }& 
       \tikz{
        \node[draw=black, line width=.5mm, inner sep=0pt] 
            {\includegraphics[width=.25\linewidth]{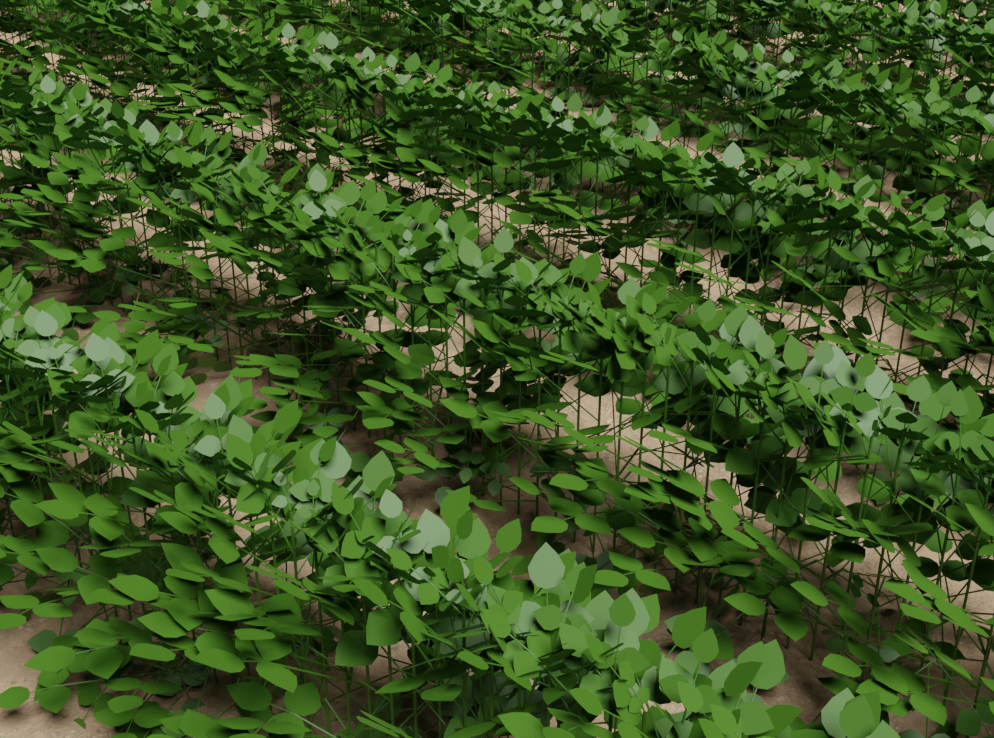}};
            ;
        } \\

        \tikz{
        \node[draw=black, line width=.5mm, inner sep=0pt] 
            {\includegraphics[width=.25\linewidth]{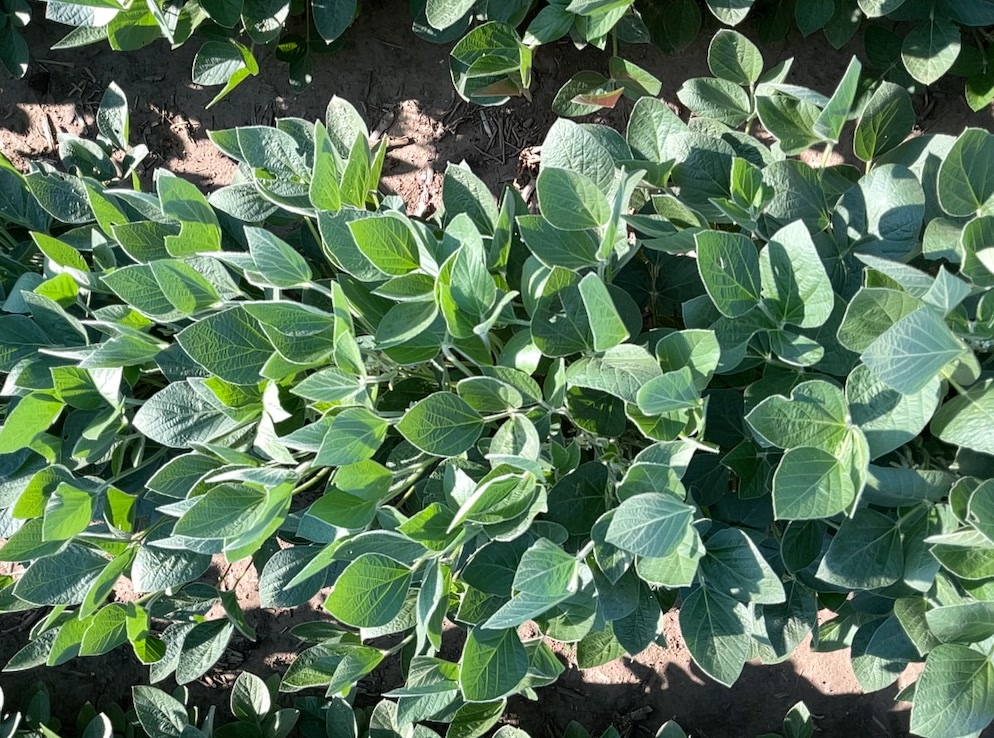}};
            \node[draw=black, draw opacity=1.0, line width=.3mm, fill opacity=0.8,fill=white, text opacity=1] at (-1.25 , 1.25) { August 01 };
        } &
        \tikz{
        \node[draw=black, line width=.5mm, inner sep=0pt] 
            {\includegraphics[width=.25\linewidth]{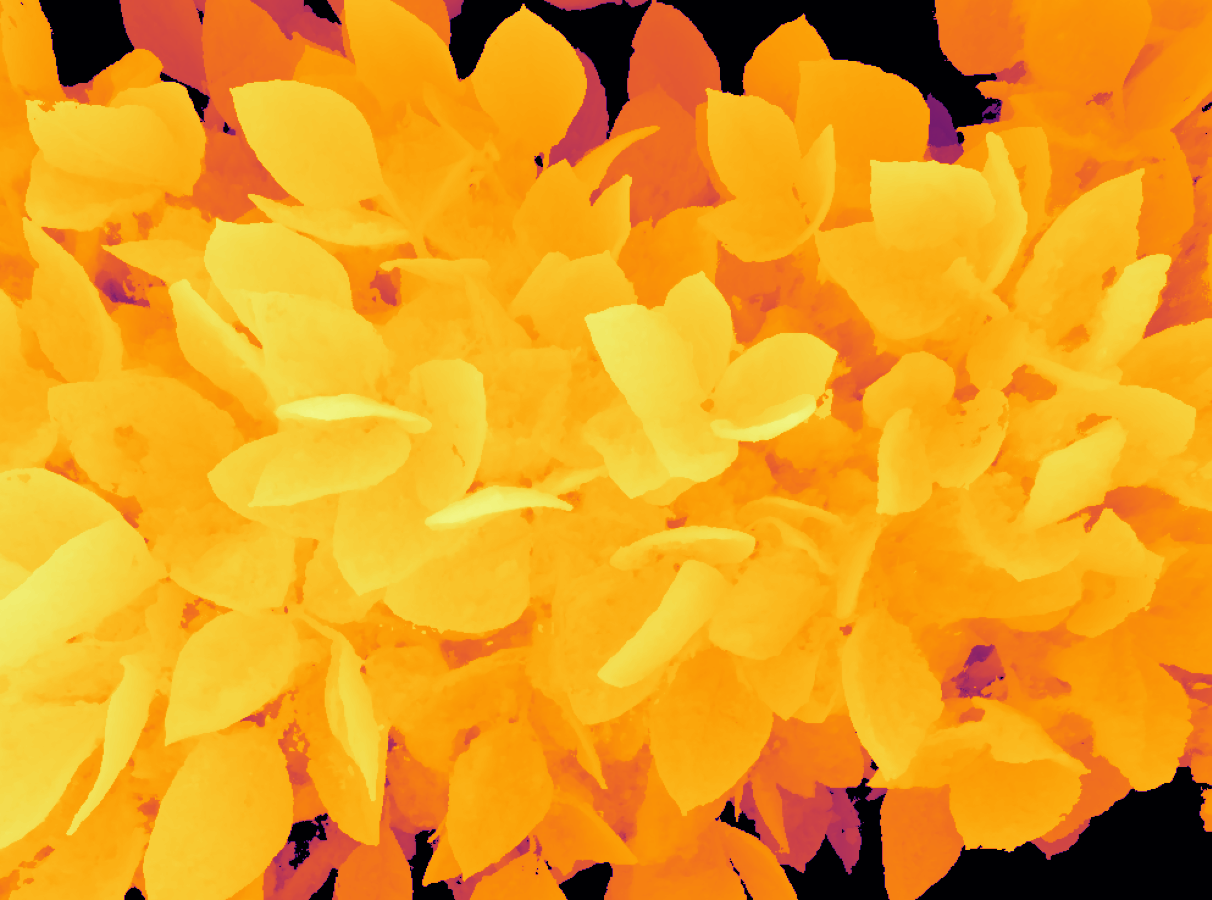}};
            ;
        }& 
        \tikz{
        \node[draw=black, line width=.5mm, inner sep=0pt] 
            {\includegraphics[width=.25\linewidth]{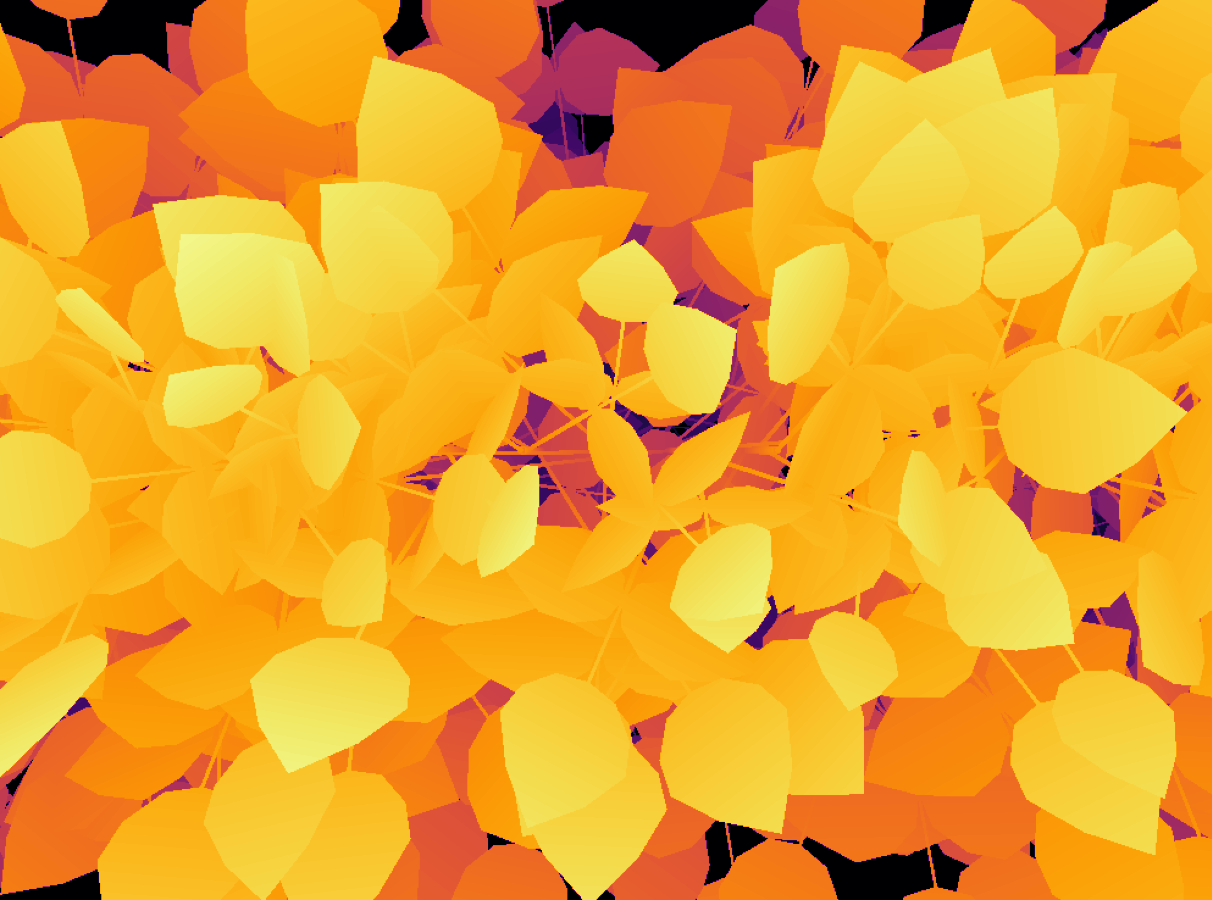}};
            ;
        }& 
       \tikz{
        \node[draw=black, line width=.5mm, inner sep=0pt] 
            {\includegraphics[width=.25\linewidth]{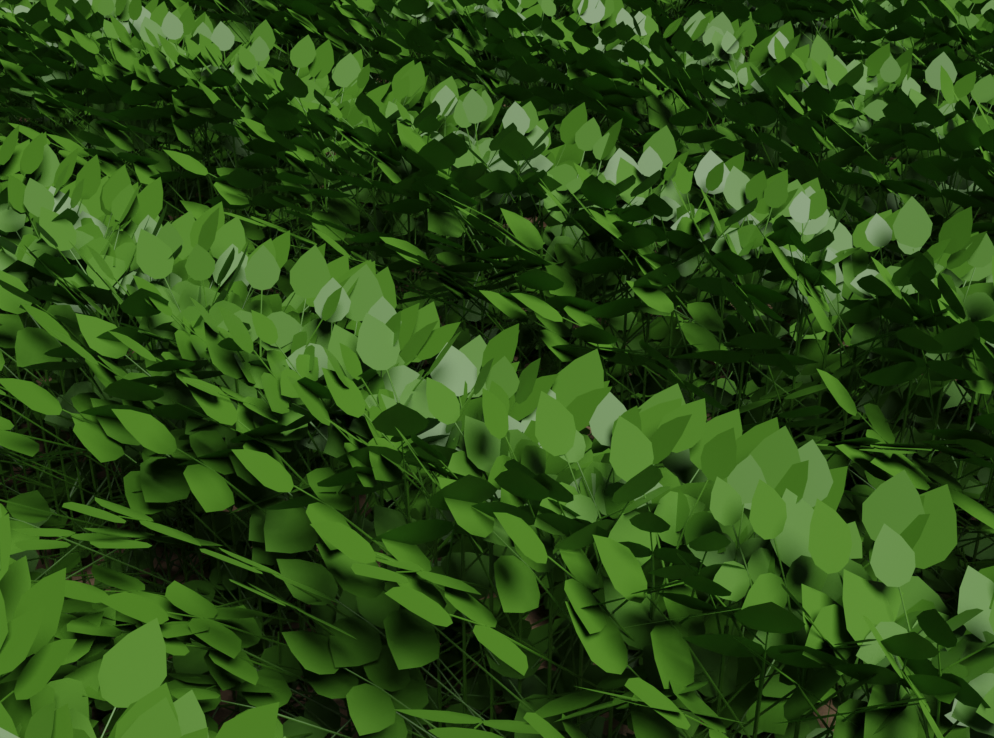}};
            ;
        } \\
       \vspace{-16pt} 
    \end{tabular}
    
    }
\captionof{figure}{\textbf{Qualitative reconstruction results (soybean)}. We validate the reconstruction quality of our method on images from real agricultural fields. From left to right: example images from the multi-view data, row-aligned NeRF-rendered depth, procedural model depth, rendered visualization of the procedural mesh. By matching statistics of the depth, our method is able to estimate the key shape parameters needed to characterize the growth of the plants throughout the growing season, consistently producing realistic reconstructions.
}
\label{fig:qual_results}
\vspace{-8pt}
\end{table*}

\begin{table*}[!t]
    \centering
    \resizebox{\linewidth}{!}{
\setlength{\tabcolsep}{0.2em} %
\renewcommand{\arraystretch}{1.}
    \begin{tabular}{cccc}
    Input & Observation Depth & Predicted Depth & 3D Mesh \\
          \tikz{
        \node[draw=black, line width=.5mm, inner sep=0pt] 
            {\includegraphics[width=.25\linewidth]{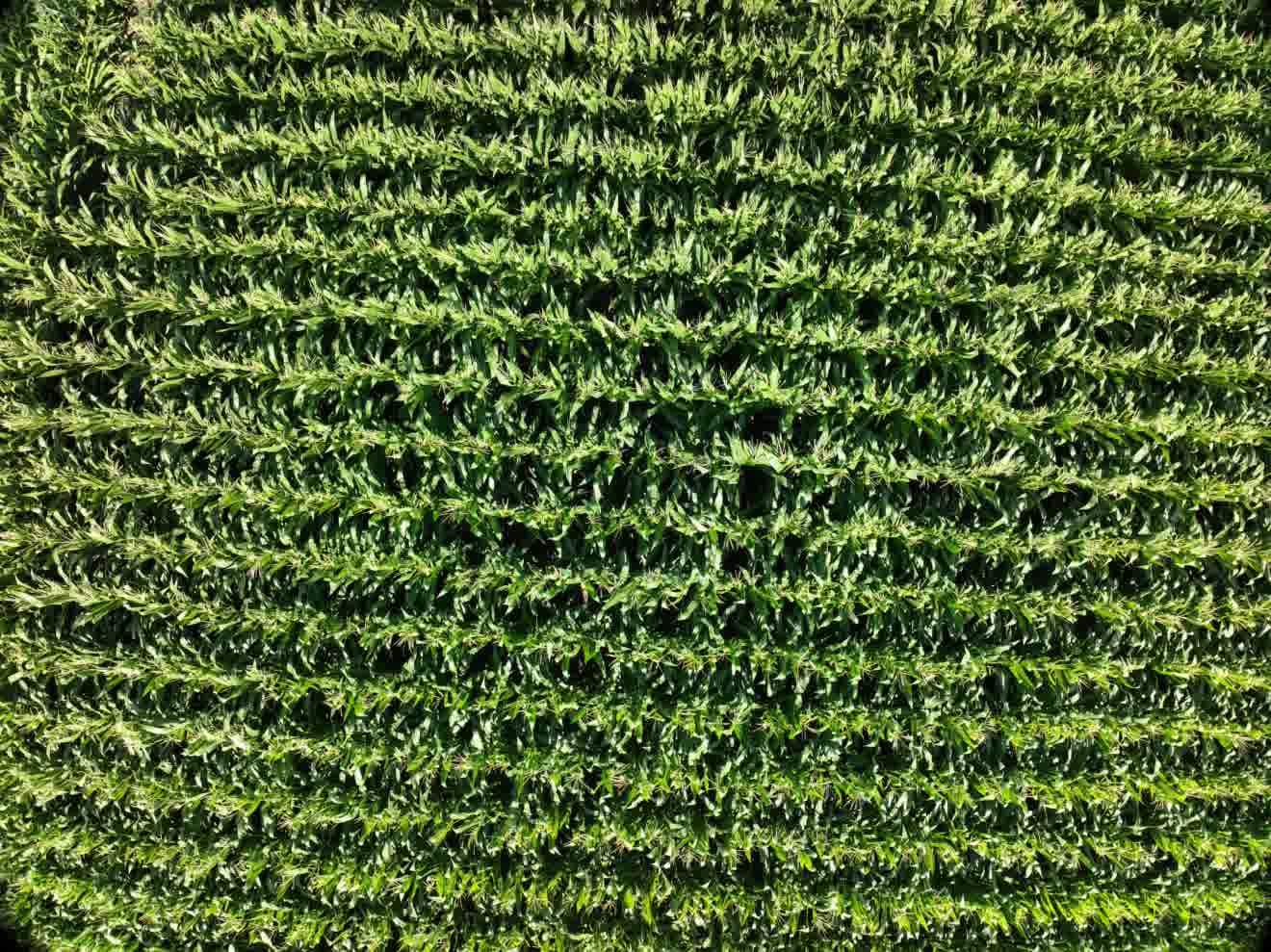}};
            \node[draw=black, draw opacity=1.0, line width=.3mm, fill opacity=0.8,fill=white, text opacity=1] at (-1.25 , 1.25) { August 16 };
        } &
        \tikz{
        \node[draw=black, line width=.5mm, inner sep=0pt] 
            {\includegraphics[width=.25\linewidth]{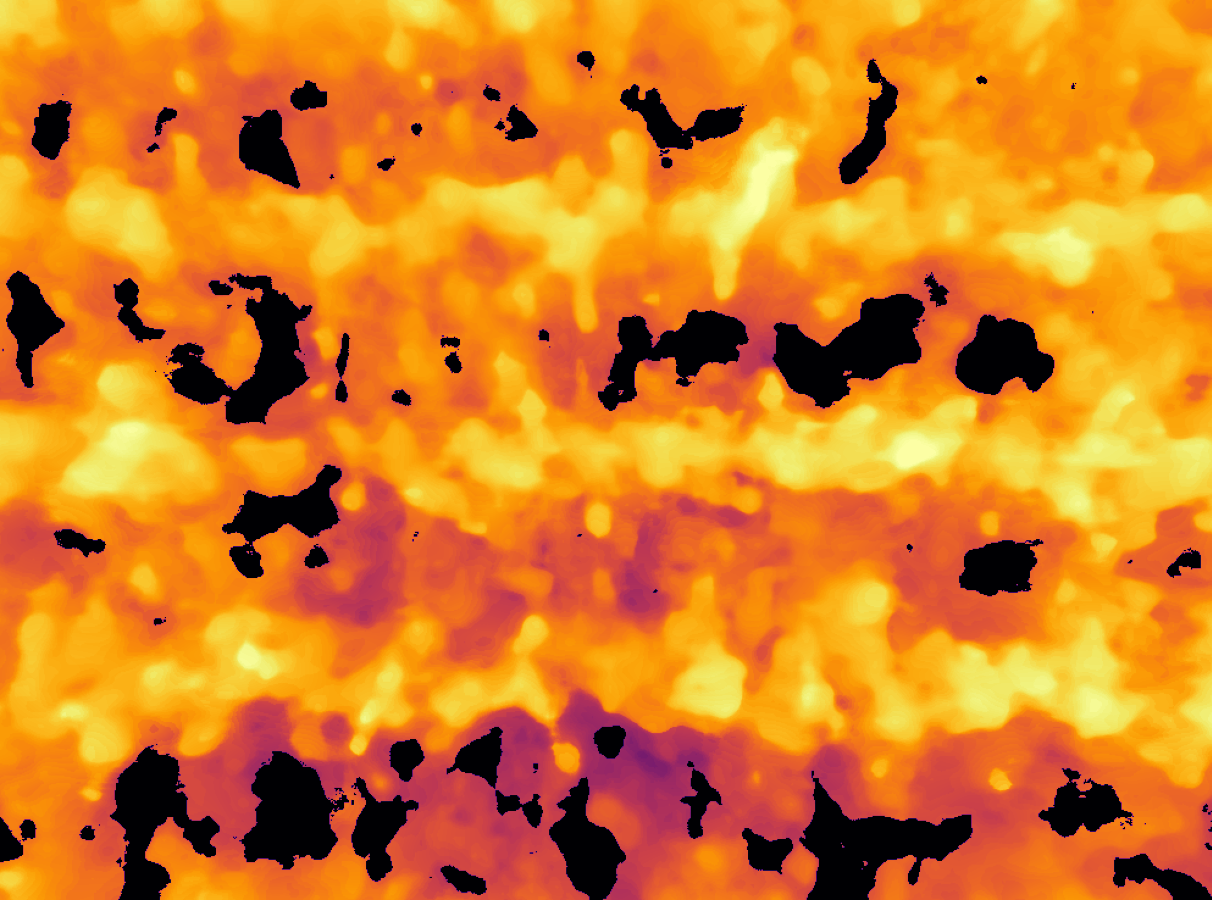}};
            ;
        }& 
        \tikz{
        \node[draw=black, line width=.5mm, inner sep=0pt] 
            {\includegraphics[width=.25\linewidth]{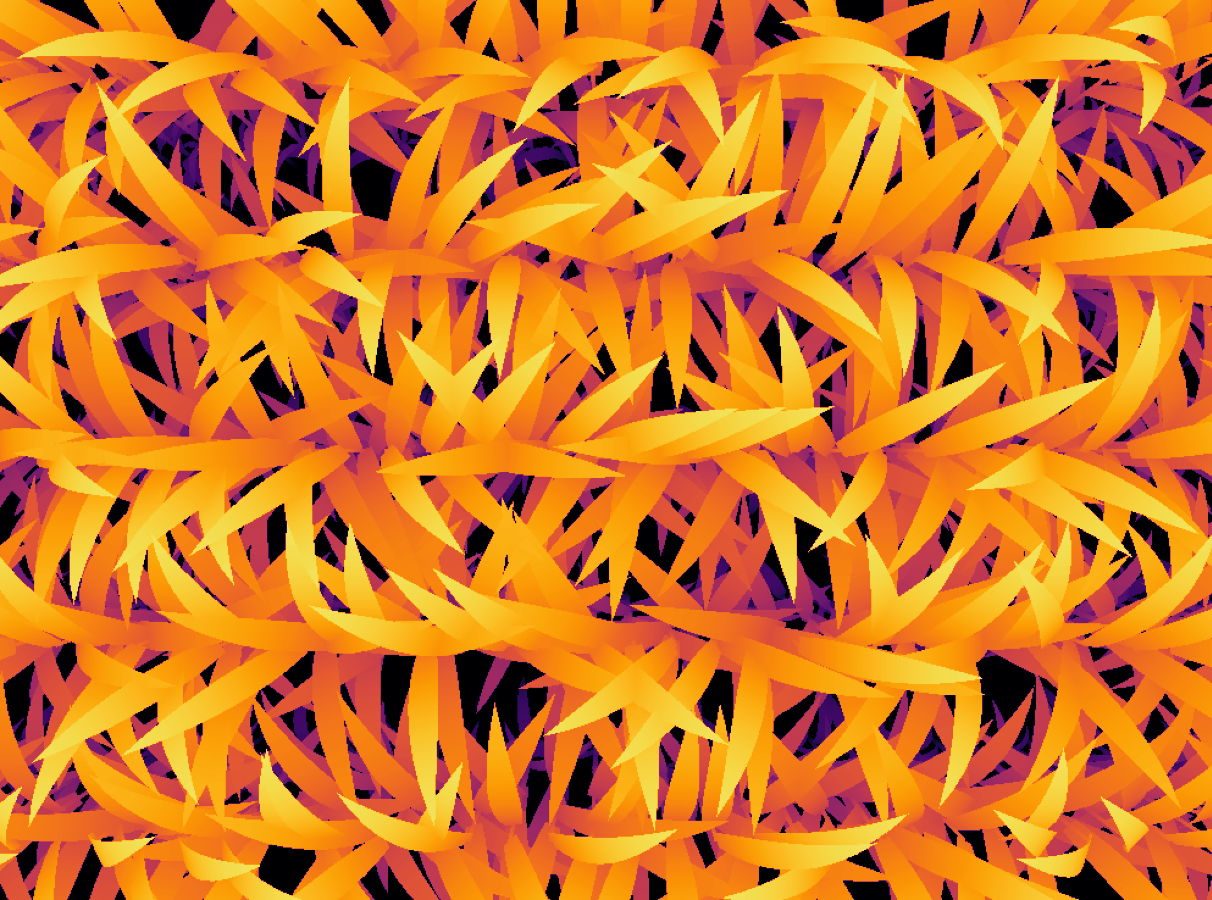}};
            ;
        }& 
       \tikz{
        \node[draw=black, line width=.5mm, inner sep=0pt] 
            {\includegraphics[width=.25\linewidth]{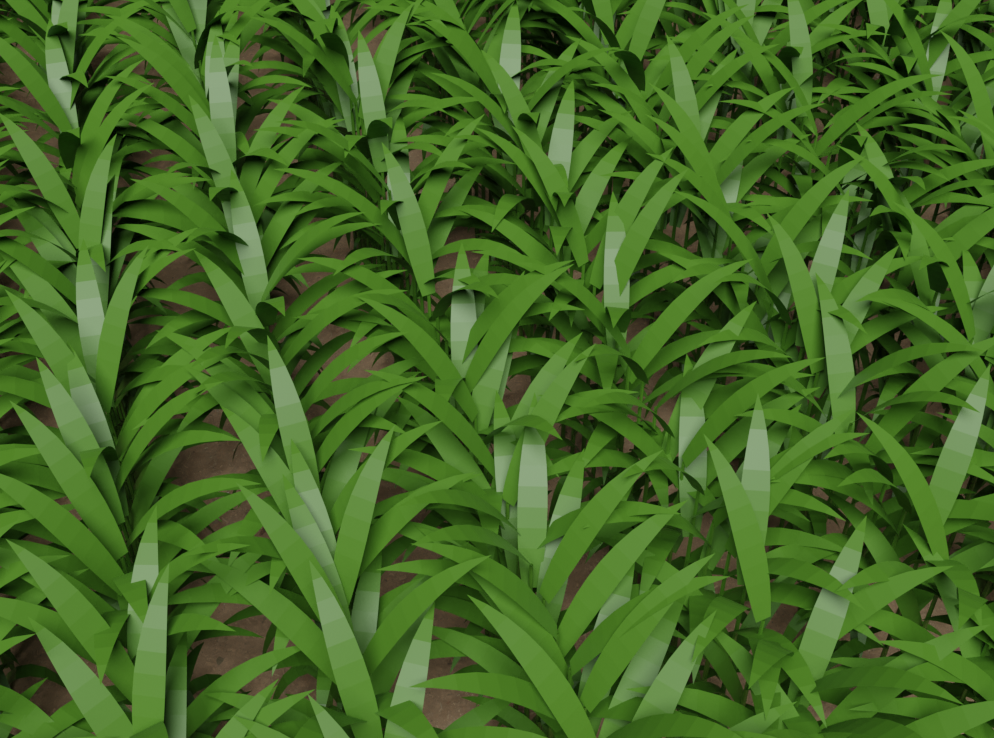}};
            ;
        } \\
        
         \tikz{
        \node[draw=black, line width=.5mm, inner sep=0pt] 
            {\includegraphics[width=.25\linewidth]{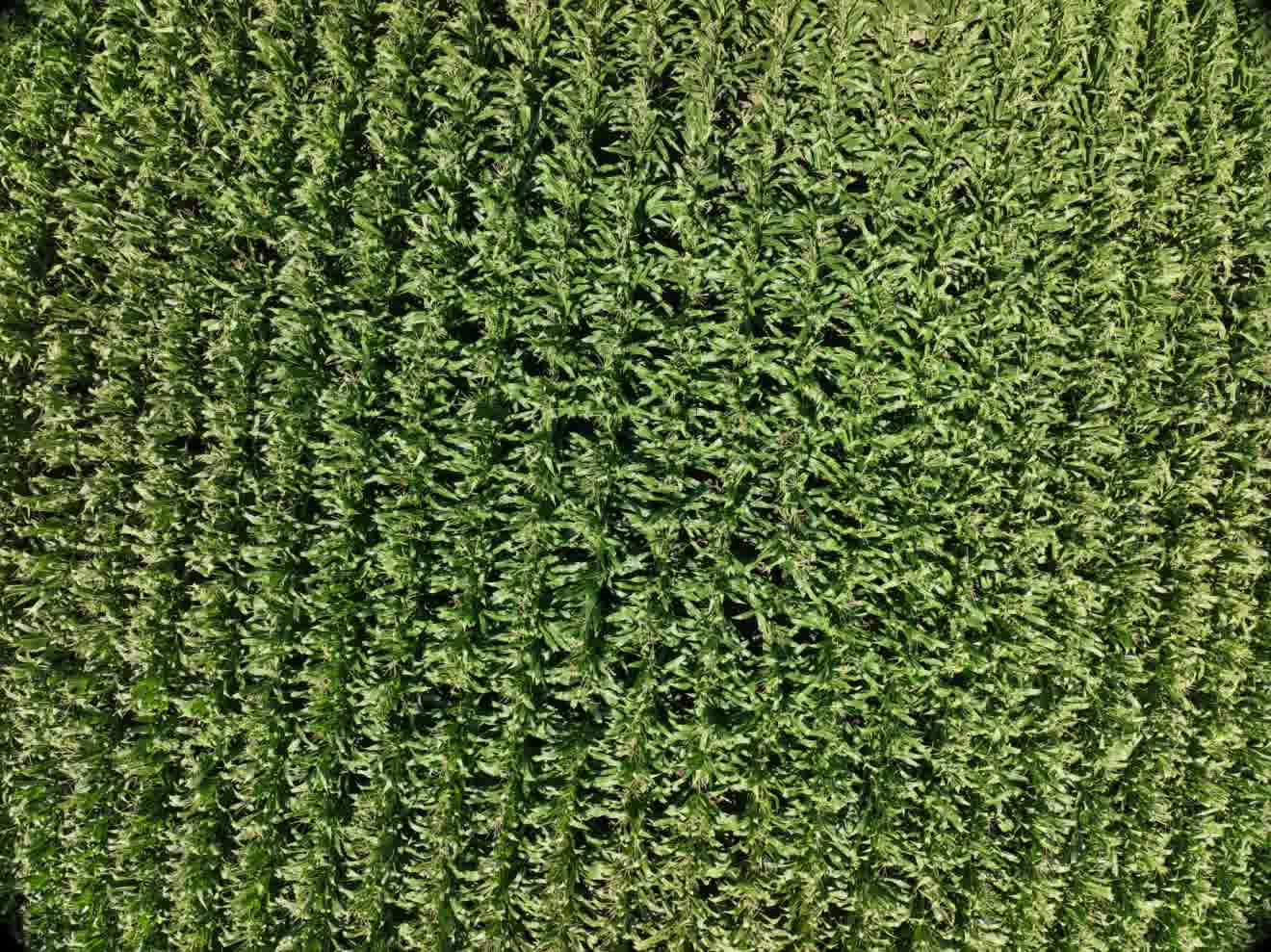}};
            \node[draw=black, draw opacity=1.0, line width=.3mm, fill opacity=0.8,fill=white, text opacity=1] at (-1.25 , 1.25) { August 28 };
        } &
        \tikz{
        \node[draw=black, line width=.5mm, inner sep=0pt] 
            {\includegraphics[width=.25\linewidth]{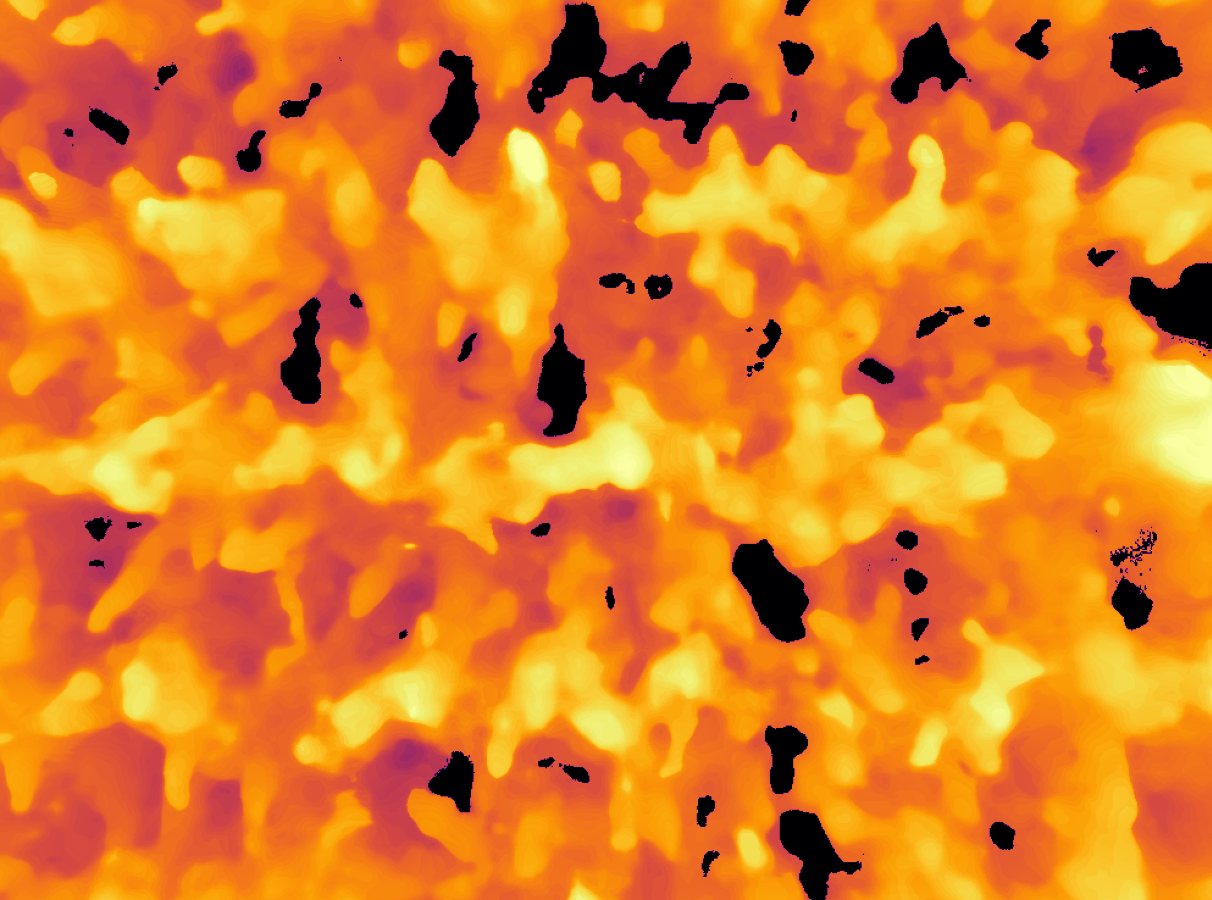}};
            ;
        }& 
        \tikz{
        \node[draw=black, line width=.5mm, inner sep=0pt] 
            {\includegraphics[width=.25\linewidth]{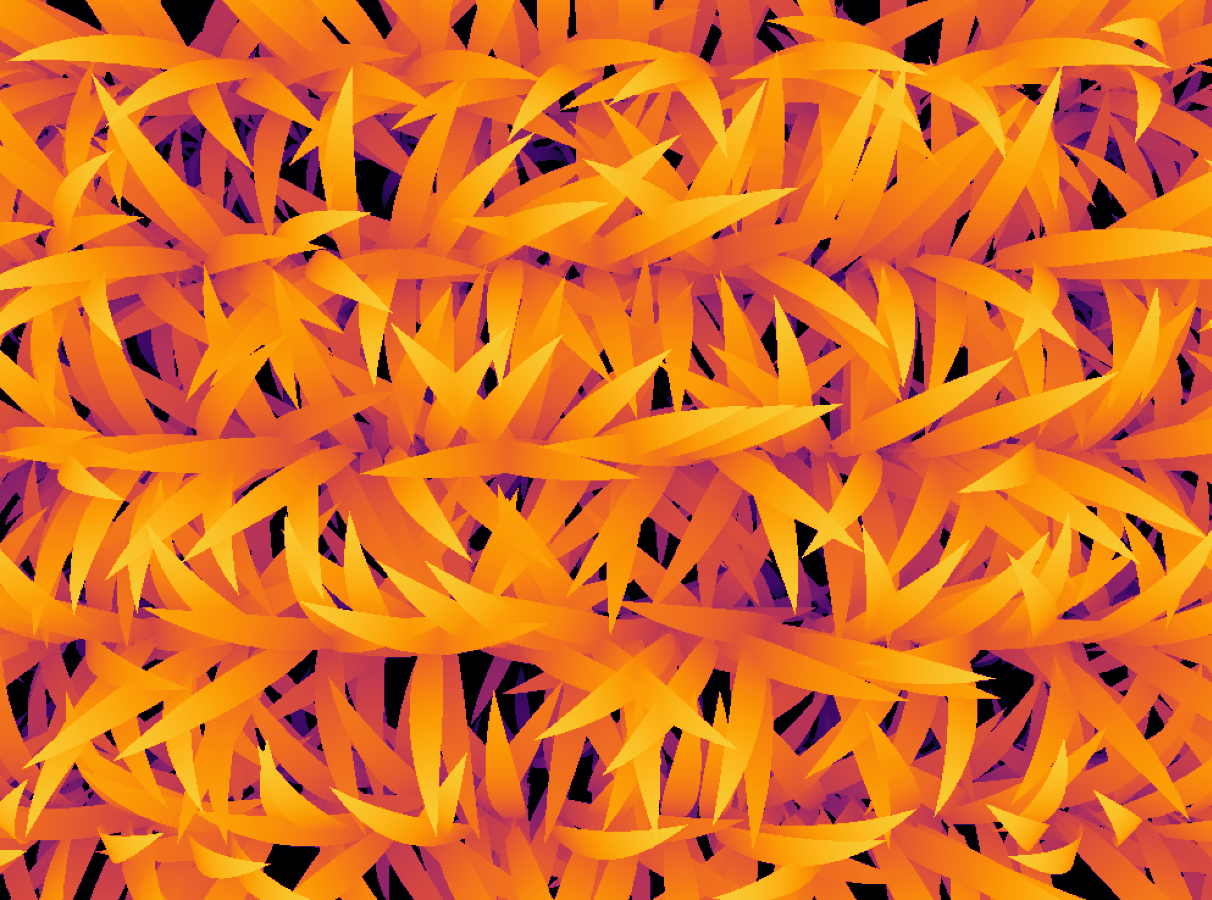}};
            ;
        }& 
       \tikz{
        \node[draw=black, line width=.5mm, inner sep=0pt] 
            {\includegraphics[width=.25\linewidth]{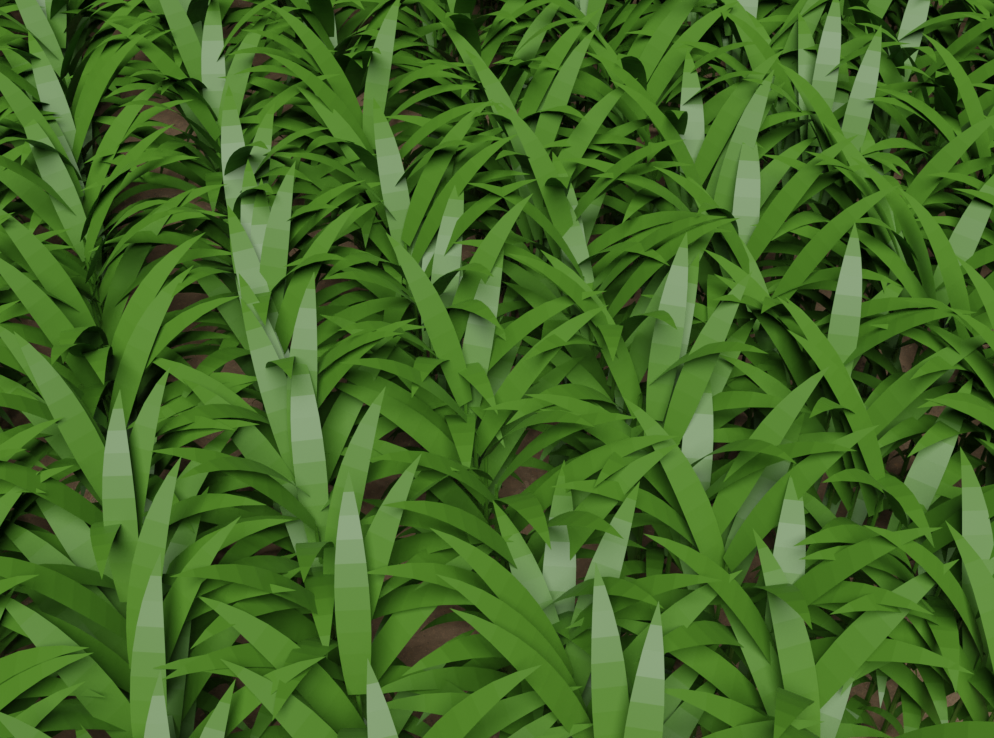}};
            ;
        } \\

       \vspace{-18pt} 
    \end{tabular}
    
    }
\captionof{figure}{\textbf{Qualitative reconstruction results (maize)}. We show that our method can be applied to model maize as well as soybean. From left to right: example images from the drone-captured multi-view data, row-aligned NeRF-rendered depth, procedural model depth, rendered visualization of the procedural mesh. The resulting reconstructions are complete and anatomically realistic despite heavy occlusion.
}
\label{fig:qual_results_maize}
\vspace{-10pt}
\end{table*}

\subsection{Loss Function}
\label{sec:loss}
One of the key motivations for estimating the 3D structure of agricultural crops is to enable photosynthesis simulations that provide accurate predictions of primary productivity and other ecological variables. Recall that we are interested in these variables at a large scale, at the level of patches or fields, not at the level of individual plants. Thus, the reconstruction of individual leaf locations is not necessary as long as certain statistics of the aggregated crop canopy are accurate. With this in mind, we design a loss function around histogram statistics of the rendered depth maps, aiming to recover the key shape properties necessary for photosynthesis simulation while avoiding overfitting to irrelevant aspects of the scene geometry.

\vspace{-6pt}
\paragraph{Depth Profile Term.} The first term of the loss function is the L2 distance between histograms of foreground depth values in the observed and predicted depth maps. Formally,
\begin{equation}
    L_{\text{depth}} = ||\mathbf{h}_{\text{obs}}  - \mathbf{h}_{\text{pred}} ||_2^2,
\end{equation}
where $\mathbf{h}_{\text{obs}}$ and $\mathbf{h}_{\text{pred}}$ are histograms of the foreground depth values in $\mathbf{I}_{\text{obs}}$ and $\mathbf{I}_{\text{pred}}$ respectively.
Unlike the depth maps themselves, the histograms are invariant to translation along the ground plane, while still containing information about the crop height. 

\vspace{-6pt}
\paragraph{Lateral Profile Term.} The lateral profile term $L_{\text{lateral}}$ is essentially the same as the depth profile term but for $y$-coordinate absolute values instead of $z$-coordinates.
This term helps to constrain the extent to which the canopy spreads out on the ground, which the depth profile is mostly invariant to.
We do not include another term for the $x$-coordinate because the $x$-axis is aligned with the planting rows, making the distribution close to uniform.

\vspace{-6pt}
\paragraph{Depth Derivative Term.} To capture more information about surface normals, we include another L2 loss $L_{\text{sobel}}$ on (histograms of) magnitudes of Sobel derivatives of the depth maps.
We sum absolute values of the Sobel derivatives in both directions with a kernel size of 3. 

\vspace{-6pt}
\paragraph{Mask Area Term.}
The final term is a squared-error on the foreground mask area:
\begin{equation}
    L_{\text{mask}} = (||\mathbf{M}_{\text{obs}}||_1 - ||\mathbf{M}_{\text{pred}}||_1)^2.
\end{equation}
Our final loss is a weighted combination of the above terms:
\begin{equation}
    L = L_{\text{depth}} + \lambda_{\text{lateral}}L_{\text{lateral}} + \lambda_{\text{sobel}}L_{\text{sobel}} + \lambda_{\text{mask}}L_{\text{mask}}.
\end{equation}

\subsection{Bayesian Optimization}
\label{sec:opt}
Since the procedural generation model directly adds new mesh faces to create plants with different topology, the transformation from parameters to generated shape is not differentiable. Thus, it is difficult to minimize the loss function with respect to the parameters using gradient-based optimization methods. Instead, we employ Bayesian optimization, a black-box optimization method that is commonly used for hyperparameter tuning~\cite{shahriari2015taking}. Bayesian optimization creates a surrogate for the objective function using Gaussian process regression, and then decides where to sample next by optimizing an acquisition function that combines uncertainty and the expected objective value~\cite{frazier2018tutorial}. We perform Bayesian optimization with a Matern kernel and the simple expected-improvement (EI) acquisition function~\cite{frazier2018tutorial} to estimate the procedural generation parameters that minimize our loss function. We find that, in many scenes, there may be multiple solutions with similar loss values. In order to make our method more robust to different random initialization seeds, we run the optimization 10 times (in parallel) and average the solutions, although this is not strictly necessary. An analysis of the distribution of the optimization results is provided in the supplementary.

\section{Experiments}
\label{sec:experiments}

\paragraph{Dataset.} To validate our approach, we collected a dataset of multi-view images in real crop fields in the U.S. Midwest, paired with manual measurements of key morphological variables. For soybean, we collected at 3 geographical locations at 6 different time points throughout the growing season, for a total of 18 scenes. The images were captured covering a \SI{2}{\meter}$\times$\SI{2}{\meter} area using the Polycam app on an iPad Pro, around 50 per scene. The paired manual measurements consist of leaf areas and leaf angles (angle between surface normal and zenith direction). Leaf area was measured using a LAI-2200 scanner, and leaf angle was measured using a protractor. 
For maize, we collected 5 different time points in one location, with each scene containing around 500 images from a DJI Mavic UAV flying at a height of \SI{13}{\meter}. The images were aligned using Agisoft Metashape. Here, only leaf area was measured, not leaf angles. Both the soybean and maize data are available through our project page.

\vspace{-8pt}
\paragraph{Metrics.} Our evaluation metrics (and field measurements) are centered around the commonly considered canopy structure variables of leaf area index (LAI) and the leaf angle distribution. LAI is defined to be the total surface area of leaves per unit of ground area and is widely used in productivity models, climate models, and methods to estimate other vegetative surface properties~\cite{parker2020tamm}. Similarly, leaf angle is regarded as a key component of plant ecological strategy with significant impact on land surface properties such as carbon flux, surface temperature and spectral signature~\cite{yang2023leaf}. To assess how accurately our reconstructions capture these key traits, we consider the following metrics: \textbf{LAI Error (LAIE)} is the root mean squared error (RMSE) between the predicted and ground-truth (manually measured) LAI. \textbf{LAI Percent Error (LAIPE)} is the absolute percent error between the predicted and ground-truth LAI. \textbf{Angle Mean Error (AME)} is the RMSE between the mean of the predicted leaf angles and the mean of the ground-truth leaf angles. \textbf{Angle Standard Deviation Error (ASDE)} is the RMSE between the standard deviation of the predicted leaf angles and that of the ground-truth leaf angles. We also evaluate the accuracy of the estimated stem structure and topology in the supplementary, using synthetic scenes instead of real scenes due to the difficulty of obtaining corresponding ground-truth measurements.

\vspace{-8pt}
\paragraph{Baselines.} We compare our reconstruction method with a variety of baselines: 
\textbf{Poisson} refers to applying Poisson surface reconstruction~\cite{kazhdan2006poisson} on the point cloud extracted from NeRF, and then thresholding in 3D to remove the background geometry. All of the remaining mesh faces are then considered as leaf faces. 
\textbf{MLP} refers to using a learned multilayer perceptron (MLP) to predict procedural generation parameters given the histograms, mask area, and view height of the observation depth map. The MLP is trained on 20K pairs of histogram statistics and model parameters, synthesized by running the procedural generation model and rendering depth maps. 
\textbf{Trust-Region} is the same as our proposed method but uses a (constrained) trust-region method~\cite{conn2000trust} for optimization instead of Bayesian optimization. This requires an estimate of the gradient of the loss function, which is obtained via 2-point finite difference estimation~\cite{strikwerda2004finite}.
 \textbf{Random} refers to uniformly randomly sampling values for the procedural generation parameters.

\vspace{-8pt}
\paragraph{Implementation Details.} For soybean, we train Nerfacto~\cite{tancik2023nerfstudio} with default parameters for 20K iterations. For maize, we set the near-plane and far-plane to 0.05 and 6.0 respectively, the MLP width to 128, and the orientation method to ``vertical''. We defer details about the RANSAC row-fitting and depth rendering to the supplementary. For our loss function, we use $\lambda_{\text{lateral}} = 2, \lambda_{\text{sobel}} = 4, \lambda_{\text{mask}} = 1$ for soybean and $\lambda_{\text{lateral}} = 1, \lambda_{\text{sobel}} = 0, \lambda_{\text{mask}} = 1$ for maize. This is because the low NeRF quality in the maize scenes makes the depth derivatives unreliable. For Bayesian optimization, we use scikit-optimize~\cite{head2020scikit} and run for 500 iterations, with 200 iterations for random initialization. Details about histogram bins and baseline implementations can be found in the supplementary.

\begin{table*}[t!]
\centering

\caption{{\bf Canopy reconstruction results.} We highlight the \colorbox{best_color}{\textbf{best}} and \colorbox{second_color}{second best} values. }
\vspace{-6pt}
{
\small
        \begin{tabularx}{0.8\textwidth}{l|>{\centering\arraybackslash}X>{\centering\arraybackslash}X>{\centering\arraybackslash}X>{\centering\arraybackslash}X|>{\centering\arraybackslash}X>{\centering\arraybackslash}X}
        \toprule
        & \multicolumn{4}{c|}{Soybean} & \multicolumn{2}{c}{Maize} \\
        & LAIE $(\downarrow)$ & LAIPE $(\downarrow)$ & AME $(\downarrow)$ &  ASDE $(\downarrow)$ & LAIE ($\downarrow$) & LAIPE $(\downarrow)$ \\
        \midrule
        Poisson  & 1.11 & 0.57 & 31.60 & {\cellcolor{second_color}{4.10}} & {\cellcolor{second_color}{1.63}} & {\cellcolor{second_color}{0.42}}\\
        MLP & {\cellcolor{second_color}{0.92}} & {\cellcolor{second_color}{0.24}} & 29.79 & \cellcolor{best_color}{\textbf{3.91}} & 2.79 & 0.88\\
        Trust-Region & 1.37  & 0.28 & \cellcolor{best_color}{\textbf{10.34}} & 7.10 & 1.71 & 0.53\\
        Random & 3.38 & 2.84 & 27.92 & 11.83 & 2.42 & 0.75 \\
        \midrule
        Ours & \cellcolor{best_color}{\textbf{0.69}} &\cellcolor{best_color}{\textbf{0.15 }}& {\cellcolor{second_color}{12.07}} & 7.39 & \cellcolor{best_color}{\textbf{0.97}} & \cellcolor{best_color}{\textbf{0.26}}\\
        \bottomrule
        \end{tabularx}
}
\vspace{-1mm}

\label{tab:canopy_structure}
\end{table*}

\begin{table}[h]
    \centering
    \resizebox{\linewidth}{!}{
\setlength{\tabcolsep}{0.15em} %
\renewcommand{\arraystretch}{1.}
    \begin{tabular}{cc}
          
        \includegraphics[width=.35\linewidth, ]{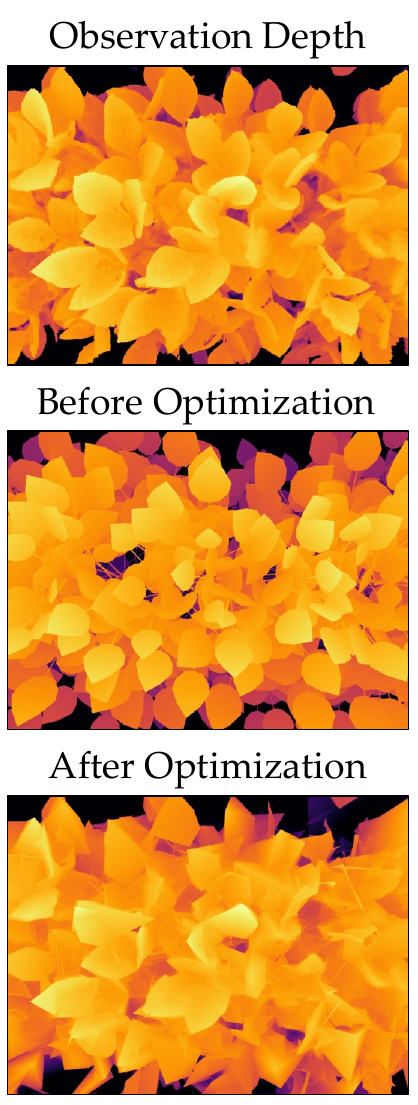}
        &
        \includegraphics[width=.65\linewidth]{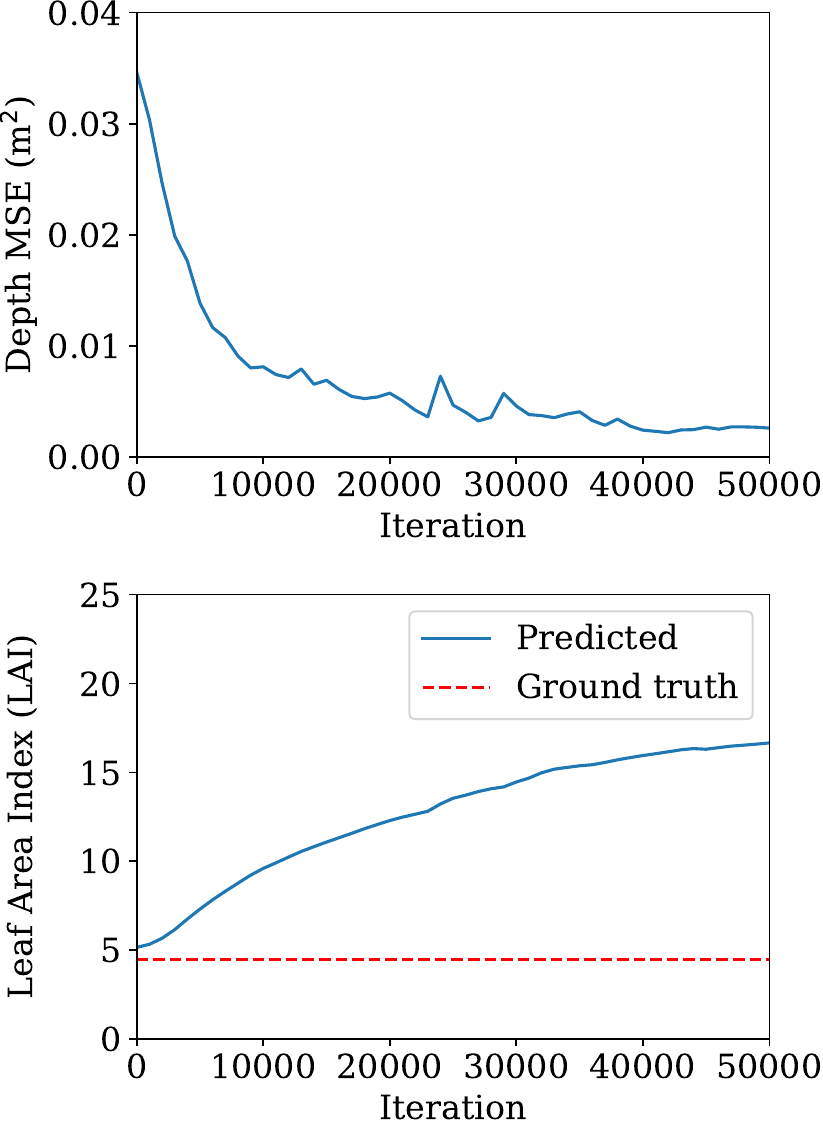}
        \\
       
       \vspace{-16pt} 
    \end{tabular}
    
    }
\captionof{figure}{\textbf{Leaf-level optimization does not improve overall structure accuracy}. We experiment with using differentiable rendering to directly optimize mesh vertices with respect to the MSE between the rendered and observed depth. Although the depth MSE decreases, the induced LAI becomes increasingly inaccurate, demonstrating the unsuitability of such vertex-level optimization. 
}
\label{fig:dr}
\vspace{-10pt}
\end{table}

\subsection{Canopy Shape Modeling}

\paragraph{Qualitative Results.}

We show example results at different stages of soybean growth for the same geographic location in Fig.~\ref{fig:qual_results}. We observe that our method produces complete reconstructions that capture how the shape of the canopy evolves over time, from small seedlings to large bushes. Unlike most existing reconstruction methods, our method is able to estimate the invisible (occluded) portions of the plants as well as the visible portions. The final mesh is also easily decomposable (at both the plant-level and part-level) and is guaranteed to be biologically reasonable thanks to the constraints imposed by the procedural model.

We also show example results for maize in Fig.~\ref{fig:qual_results_maize}. Due to windy conditions, larger scene scale, and noisy camera pose estimation, the quality of the NeRF is significantly lower in the maize data compared to soybean. However, our method is still able to produce reasonable canopy reconstructions using those observations.

\vspace{-6pt}
\paragraph{Quantitative Results.}

We provide quantitative results for the canopy reconstruction in Tab.~\ref{tab:canopy_structure}.  In all metrics except ASDE, we find that our proposed method achieves the highest performance. The angle standard deviation may be estimated less accurately than the other variables because it does not have as strong of an impact on the loss function.

The Poisson reconstruction fails because it completely ignores occluded regions and also frequently stitches nearby leaves together, creating extraneous faces. The MLP gives reasonable predictions in terms of soybean leaf area, but tends to give highly erroneous leaf angles as a result of the domain gap between the synthetic histograms it was trained on and the actual histograms from the NeRF depth maps. This is exacerbated with the maize data, where the NeRF renderings are blurrier. The trust-region optimization produces leaf angles comparable with our (Bayesian optimization) method, but does not perform as well for leaf area, most likely falling into local minima or making erroneous steps due to the stochasticity of the procedural model. 

\vspace{-6pt}
\paragraph{Naive geometric optimization.} We emphasize that our goal is \textit{not} exact leaf-level reconstruction of the crop canopy. Instead, our task demands complete crop canopies that capture key shape statistics of the actual leaf distribution, such that downstream applications like growth monitoring or photosynthesis prediction can be conducted accurately. Our model parameterizations and loss function terms are designed to capture precisely these key shape variations, and not overfit to irrelevant geometric details of the input observations. To illustrate this point, we conduct an experiment using differentiable rendering to directly optimize the vertices of our output mesh to further fit the observed depth map. The optimization is done in PyTorch3D~\cite{ravi2020accelerating} using SGD with learning rate 0.01 and momentum 0.9 on an MSE loss. The results are shown in Fig.~\ref{fig:dr}. Although the pixel-wise depth error can be made very low, important variables such as the total leaf area become \textit{less} accurate, which is reflected in the LAI~\cite{parker2020tamm}. This justifies the need for shape constraints and distribution-level optimization.

\subsection{Ablation Studies}
\begin{table*}[t!]
\centering
\caption{{\bf Loss function ablations.}  We highlight the \colorbox{best_color}{\textbf{best}} and \colorbox{second_color}{second best} values.  }
\vspace{-6pt}
{
\small

        \begin{tabularx}{0.8\textwidth}{cccc|>{\centering\arraybackslash}X>{\centering\arraybackslash}X>{\centering\arraybackslash}X>{\centering\arraybackslash}X}
        \toprule
        \multicolumn{4}{c|}{Loss Components} & \multicolumn{4}{c}{Soybean} \\
        $L_{\text{depth}}$ & $L_{\text{lateral}}$ & $L_{\text{sobel}}$ & $L_{\text{mask}}$ & LAIE $(\downarrow)$ & LAIPE $(\downarrow)$ & AME $(\downarrow)$ &  ASDE $(\downarrow)$  \\
        \midrule
          \checkmark & &  &  & 1.62 & 0.52 & 17.00 & 9.26\\
         & \checkmark & \checkmark & \checkmark &\cellcolor{best_color}{\textbf{ 0.67}} & 0.19 & 12.08  &\cellcolor{best_color}{\textbf{ 6.81} }\\
        \checkmark &  & \checkmark & \checkmark & 0.71 & {\cellcolor{second_color}{0.16}} & {\cellcolor{second_color}{10.37}} & 7.69 \\
        \checkmark & \checkmark &  & \checkmark & 1.94 & 0.49 & \cellcolor{best_color}{\textbf{9.73 }}& 8.15 \\
        \checkmark & \checkmark & \checkmark & & 0.78 & 0.19 & 12.30 & 8.03 \\
        \midrule
        \checkmark & \checkmark & \checkmark & \checkmark & {\cellcolor{second_color}{0.69}} & \cellcolor{best_color}{\textbf{0.15}} & 12.07 & {\cellcolor{second_color}{7.39}} \\
        \bottomrule
        \end{tabularx}
}
\vspace{-8pt}

\label{tab:ablation}
\end{table*}
We conduct an ablation study on the components of our loss function in Tab.~\ref{tab:ablation}. We observe that each of the components contribute to producing a more accurate reconstruction, some significantly more than others. The depth derivative loss has a pronounced beneficial effect on the leaf area metrics. This is because the other statistics may have trouble constraining the leaf size when the canopy has covered most of the ground, while the depth derivative distribution will be strongly shifted upwards with smaller leaves compared to larger ones. We note that removing some of the other terms one at a time does not actually significantly harm the final performance. The hard constraints imposed by the procedural model and the soft constraints imposed by the remaining terms are able to force a similar solution. We decide to include all the terms in our final method due to the improved LAIPE, which we prefer over LAIE as it is not biased towards the later time points with larger plants. 

We also test our method with different procedural models to investigate the importance of having realistic shape priors. The results can be found in the supplementary, along with a discussion of limitations.

\begin{table}[]
    \centering
    \resizebox{\linewidth}{!}{
\setlength{\tabcolsep}{0.1em} %
\renewcommand{\arraystretch}{1.}
    \begin{tabular}{ccc}
    \small Net Photosynthesis & \small  \small Noon (12 PM) & \small Late Afternooon (4 PM)  \\
           
        \includegraphics[width=.32\linewidth, trim={0 0 0 0.5cm}, clip]{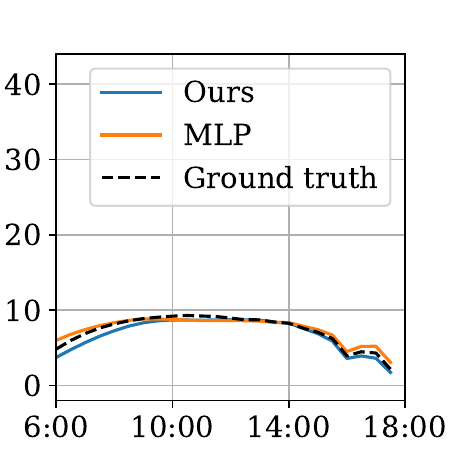}
        &
        
        \includegraphics[width=.34\linewidth]{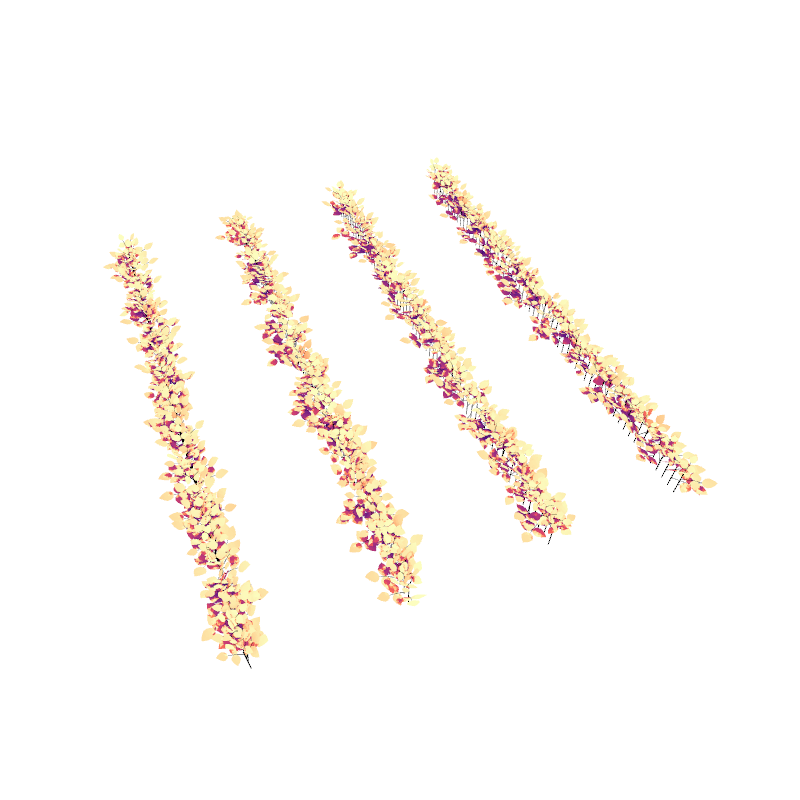}
        & 

        \includegraphics[width=.34\linewidth]{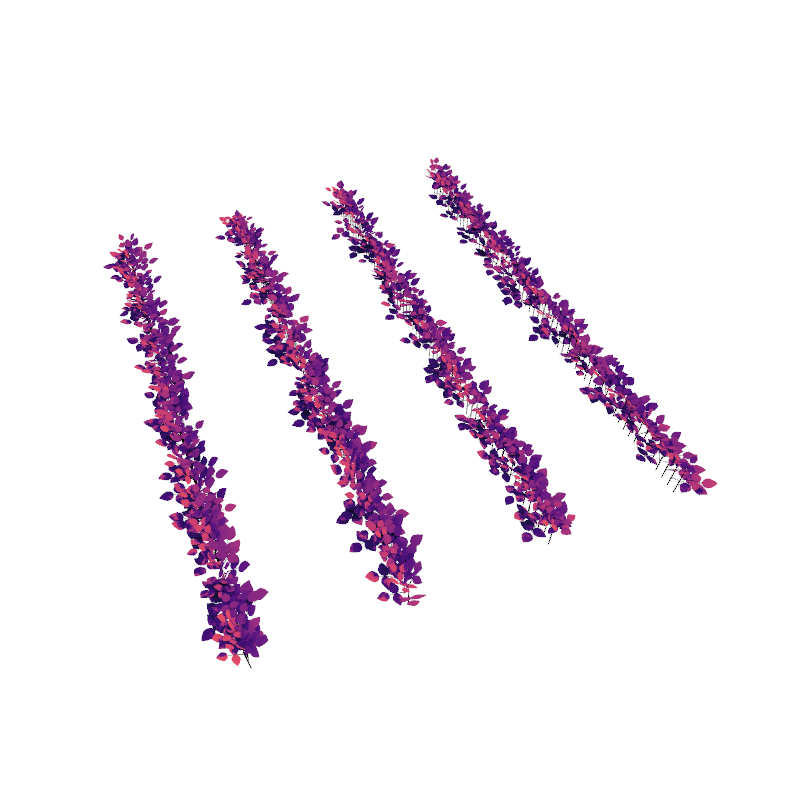}
        \vspace{-6pt} 
        \\

        \includegraphics[width=.32\linewidth]{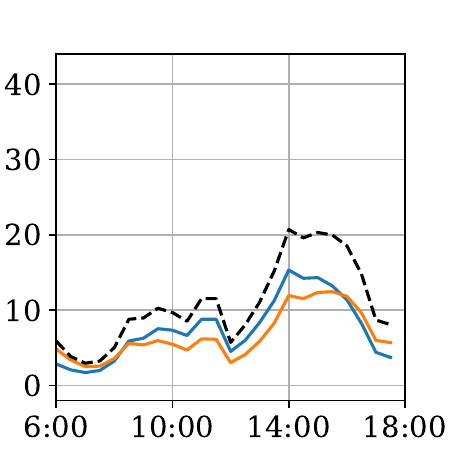}
        &
       
        \includegraphics[width=.34\linewidth]{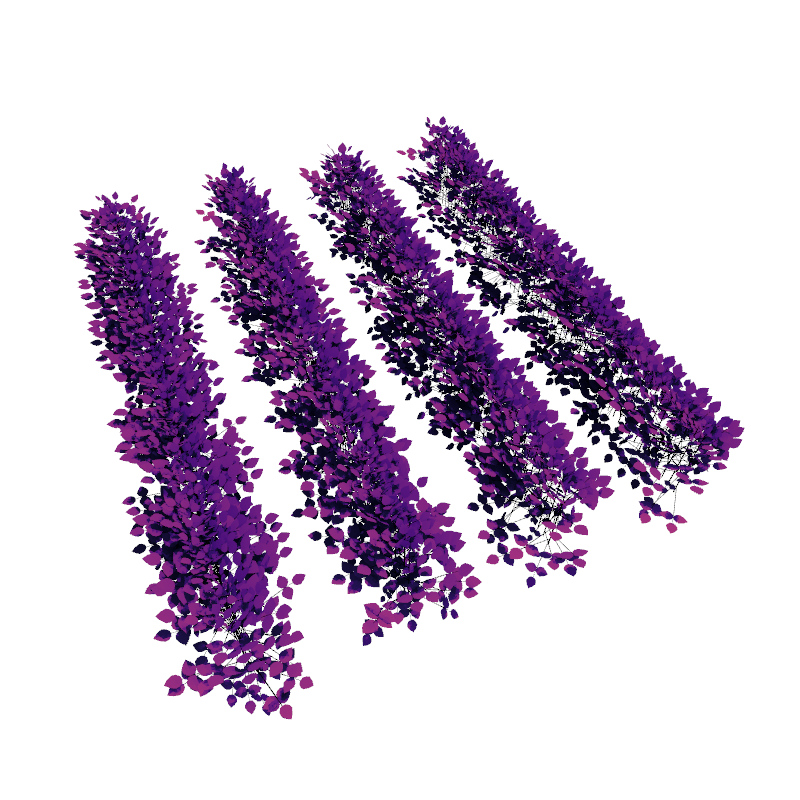}
        & 

        \includegraphics[width=.34\linewidth]{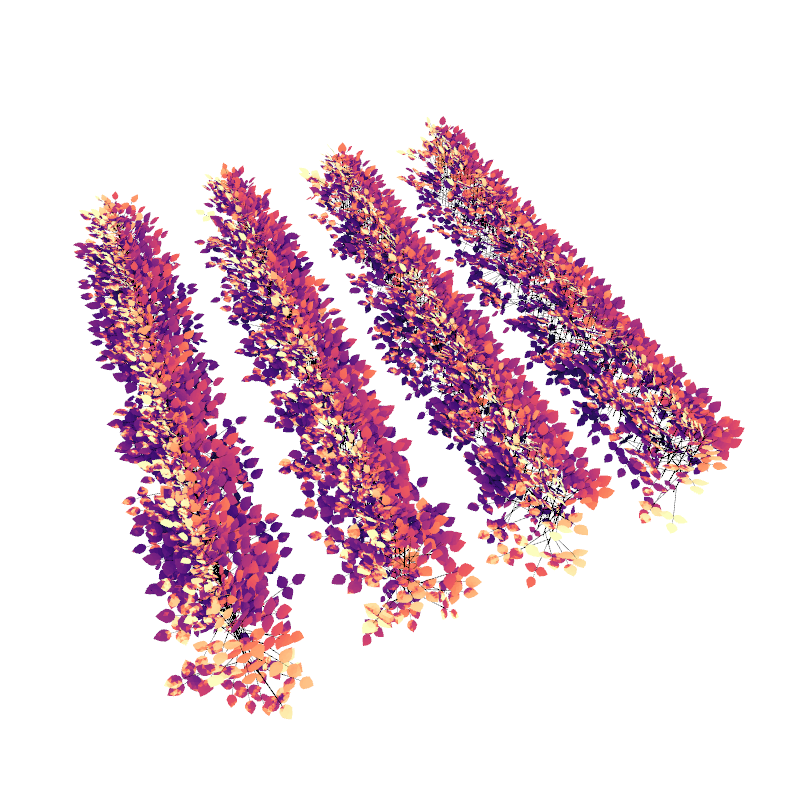}
        \vspace{-6pt} 
        \\

        \includegraphics[width=.32\linewidth]{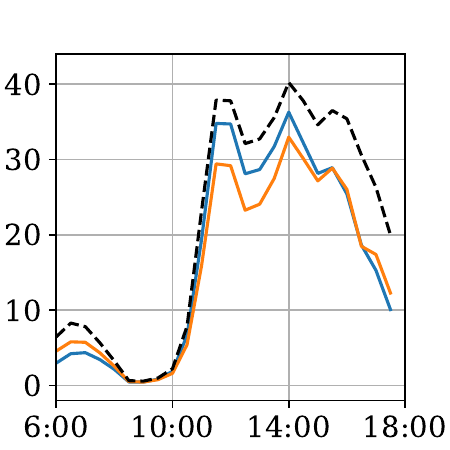}
        &
        
        \includegraphics[width=.34\linewidth]{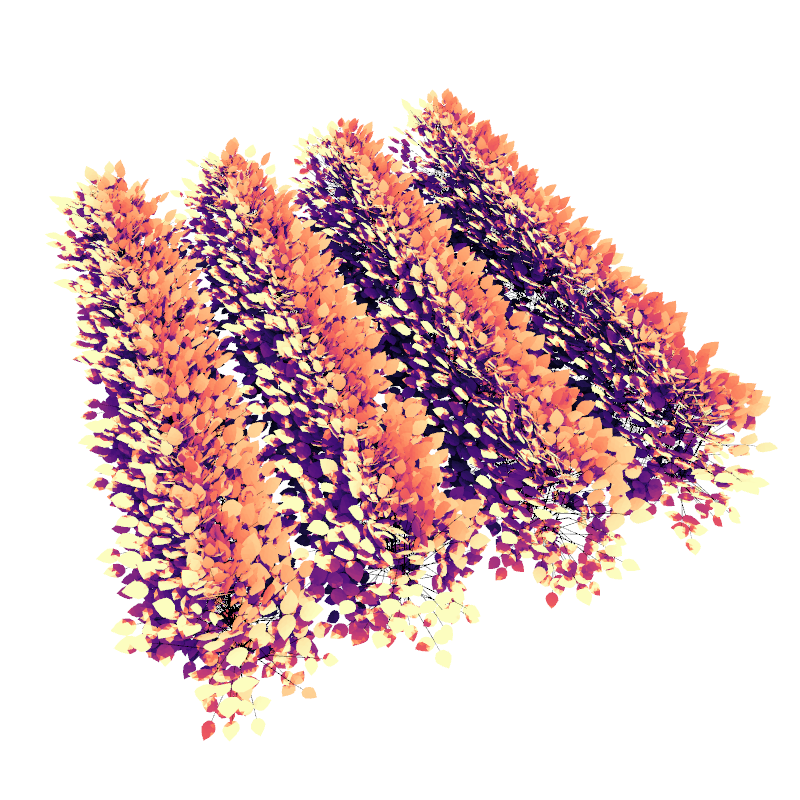}
        & 

        \includegraphics[width=.34\linewidth]{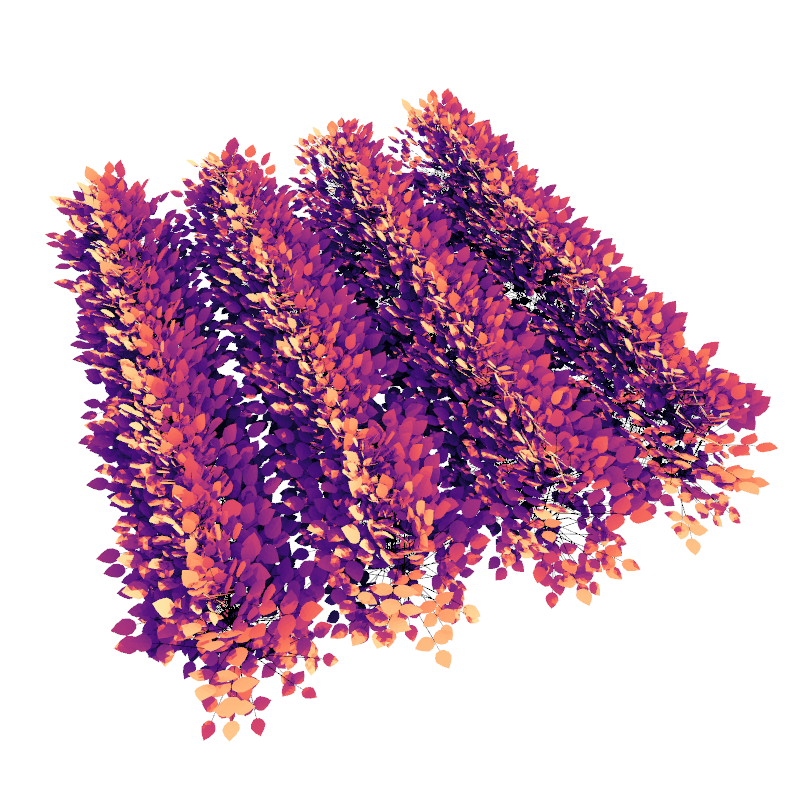}
        \vspace{-4pt} 
        \\

    \end{tabular}
    
    }
    \vspace{-6pt}
\captionof{figure}{\textbf{Photosynthesis simulation results}. We perform simulations using Helios~\cite{bailey2019helios} on soybean canopies reconstructed by our method. The left column shows a timeseries of the net photosynthesis rate for the crop canopy over the course of a day, in units of $\mu$molCO$_2$/\SI{}{\meter}$^2$/\SI{}{\second}. The other columns show a mesh visualization where leaf faces are colored by the rate of photosynthesis over that face (brighter = higher rate). The results demonstrate the potential of using our pipeline for monitoring crop productivity. 
}
\label{fig:photosynthesis}
\vspace{-12pt}
\end{table}

\subsection{Photosynthesis Simulation}

The outputs of our method can be directly used with radiative transfer models to predict photosynthesis rates, which directly impacts crop productivity. We showcase this using Helios~\cite{bailey2019helios}, a state-of-the-art biophysical modeling framework. Helios can predict the net photosynthesis rate for every leaf mesh face at any point in time given a set of environmental variables measured by a flux tower with sensors for temperature, humidity, radiation, etc. We show visualizations of the per-face photosynthesis rate as well as timeseries of the aggregate rate across the entire canopy in Fig.~\ref{fig:photosynthesis}. Ground-truth values for the photosynthesis rate are usually calculated at a landscape-level using eddy covariance data from a flux tower~\cite{pastorello2020fluxnet2015}. Since we do not have paired flux tower measurements for our dataset (collected in 2023), we use measurements from a soybean field in the previous year (2022) with a similar location and climate. 

The predictions appear to accurately represent the net photosynthesis over time as well as its spatial variations throughout the canopy, validating our pipeline's potential for facilitating large-scale monitoring of crop productivity.

\section{Conclusion}
\label{sec:conc}

We presented a novel method for 3D modeling of agricultural crops that combines neural rendering with inverse procedural modeling to produce complete plant shapes despite heavy occlusion. Our method first constructs a neural radiance field (NeRF) and then optimizes a procedural model to be consistent with the NeRF in a visibility-aware manner. We validate our method on real-world agricultural fields and show that it can reconstruct realistic crop canopies across a variety of growth stages. We also show that it enables realistic simulations of crop photosynthesis using a radiative transfer model. The method could potentially be improved in the future by incorporating plant growth priors for temporal consistency, or optimizing the model in a coarse-to-fine manner to capture more shape details.
\vspace{-16pt}
\paragraph{Acknowledgments.} This work is supported by NSF Awards \#2331878, \#2340254, \#2312102, \#2414227, and \#2404385, an Intel research gift, the NCSA Faculty Fellowship, and the UIUC Agroecosystem Sustainability Center.

{
    \small
    \bibliographystyle{ieeenat_fullname}
    \bibliography{main}
}

\renewcommand*{\thesection}{\Alph{section}}
\maketitlesupplementary

\begin{abstract}
In the following supplementary material, we provide additional experimental details (Sec.~\ref{sec:additional_details}), additional experimental results (Sec.~\ref{sec:additional results}), and a discussion of limitations (Sec.~\ref{sec:limitations}). We invite readers to watch the supplementary video (\texttt{supp.mp4}) for further visualization.
\end{abstract}

\section{Additional Experimental Details}
\label{sec:additional_details}

\subsection{Additional Procedural Model Details}
\paragraph{Soybean Model.} Our procedural soybean model is based on the mCanopy model from Song et al.~\cite{song2020decomposition}. Each plant consists of a main stem with up to 14 trifoliate nodes, and up to 6 branch stems with up to 2 nodes each. Unifoliate nodes and cotyldeons are ignored for simplicity. Each node consists of three coplanar leaves, with the middle leaf connected by a small petiole, and is connected to its stem by a larger petiole. The leaf length (and width), the petiole length, the angle with the stem, and the distance between nodes are variable, as they are affected by the multiplier parameters. On the main stem, when all the multipliers are set to 1, the variables follow a profile from top to bottom based on field data measured by Song et al.~\cite{song2020decomposition}. Branches start growing with every new main stem node starting from the 8th, with one branch per main stem node starting from the bottom and up to the 6th. The number of nodes per branch is selected randomly according to the distribution defined by Song et al.~\cite{song2020decomposition}. The variables for the branch nodes are the same across nodes (but are affected by the multipliers). When the number of nodes is not an integer, it is rounded up and the last node is scaled down in size linearly according to the remainder. Additionally, random noise is added to each petiole angle following a Gaussian distribution with mean 0 and standard deviation 5 degrees. Each node grows on alternating sides of their stem, with additional azimuthal rotation following a Gaussian distribution with mean 0 and standard deviation 60 degrees. The range of possible values for each model parameter are provided in Tab~\ref{tab:soybean_param_ranges}.

\begin{table}[h!]
\centering
{
\small
        \begin{tabular}{l|c}
        \toprule
        Parameter & Range \\
        \midrule 
        Leaf length multiplier & 0.5 - 1.5 \\
        Petiole length multiplier & 0.5 - 2.0 \\
        Petiole angle multiplier & 0.5 - 4.0 \\
        Internode length multiplier & 0.5 - 2.0 \\
        Number of nodes & 1.0 - 14.0 \\
        \bottomrule
        \end{tabular}
}
\caption{{\bf Soybean model parameter ranges.} }
\label{tab:soybean_param_ranges}
\end{table}

\paragraph{Maize Model.} Our procedural maize model is based no the coupled maize model from Qian et al.~\cite{qian2023coupled}. Each plant consists of a main stem with up to 18 nodes with one leaf each. Each leaf consists of 10 trapezoidal segments that bend towards the ground following a polynomial function that depends on the leaf order~\cite{qian2023coupled}. To allow for different bending angles, we introduce a parameter that additively shifts all of the leaf orders by the same amount. The leaf lengths, widths, and internode lengths follow a profile from bottom to top based on field data measured in the U.S. Midwest. Each node grows on alternating sides of their stem, with additional azimuthal rotation following a Gaussian distribution with mean 0 and standard deviation 60 degrees. The range of possible values for each model parameter are provided in Tab~\ref{tab:maize_param_ranges}.

\begin{table}[h!]
\centering
{
\small
        \begin{tabular}{l|c}
        \toprule
        Parameter & Range \\
        \midrule 
        Leaf length multiplier & 0.8 - 1.2 \\
        Leaf order shift & -4.0 - 4.0 \\
        Internode length multiplier & 0.8 - 1.2 \\
        Number of nodes & 1.0 - 18.0 \\
        \bottomrule
        \end{tabular}
}
\caption{{\bf Maize model parameter ranges.} }
\label{tab:maize_param_ranges}
\end{table}

\begin{table*}[!t]
    \centering
    \resizebox{\linewidth}{!}{
\setlength{\tabcolsep}{0.2em} %
\renewcommand{\arraystretch}{1.}
    \begin{tabular}{ccccc}
    Input & Poisson & MLP & Trust-Region & Ours \\
          \tikz{
        \node[draw=black, line width=.5mm, inner sep=0pt] 
            {\includegraphics[width=.25\linewidth]{fig/qual_results/input0627.jpg}};
            \node[draw=black, draw opacity=1.0, line width=.3mm, fill opacity=0.8,fill=white, text opacity=1] at (-1.4 , 1.25) { June 27 };
        } &
        \includegraphics[width=.19\linewidth]{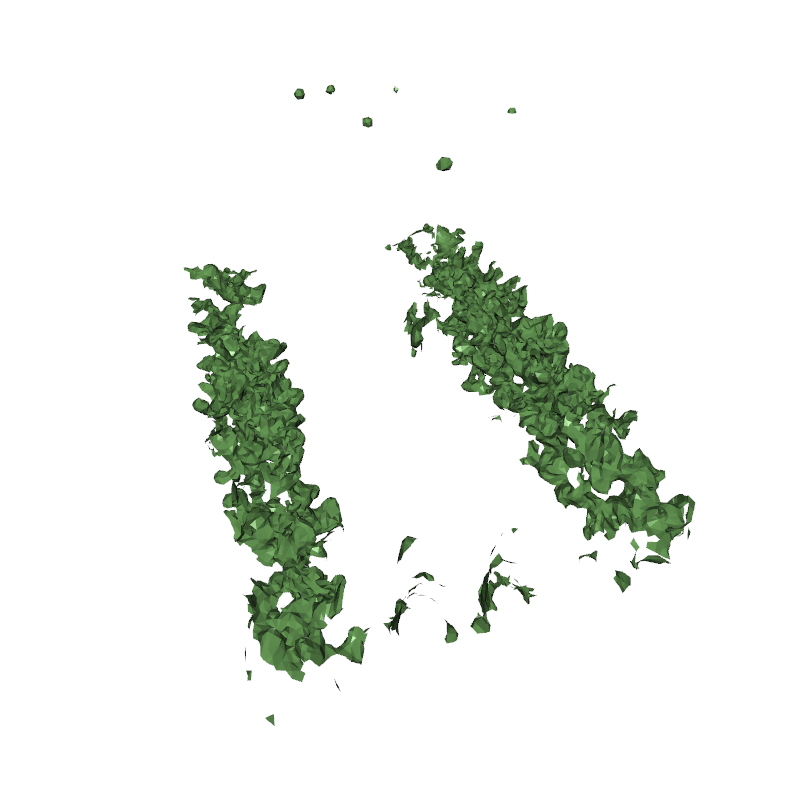}& 
        \includegraphics[width=.19\linewidth]{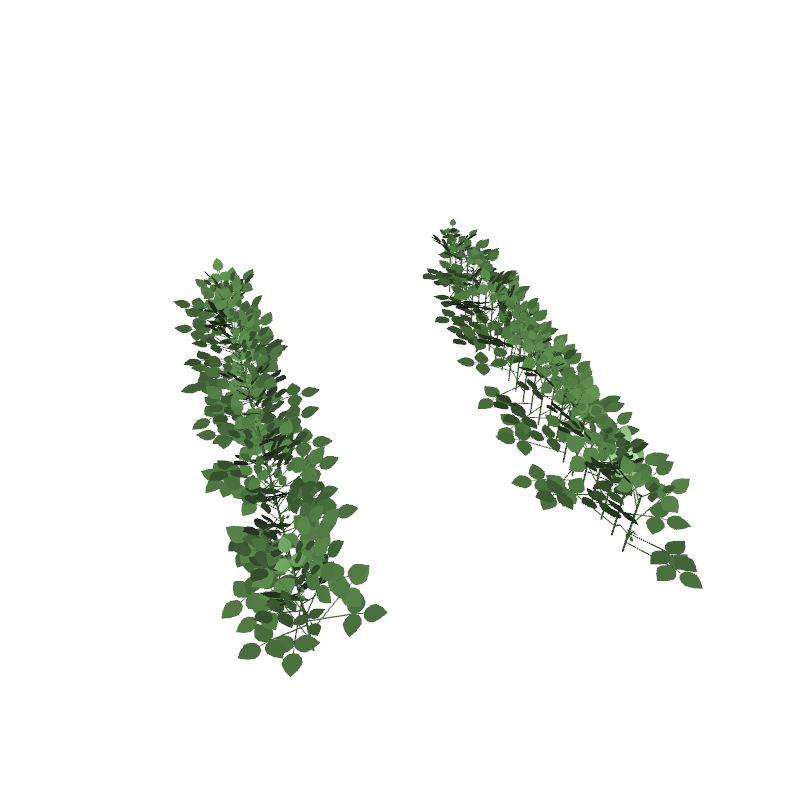}& 
        \includegraphics[width=.19\linewidth]{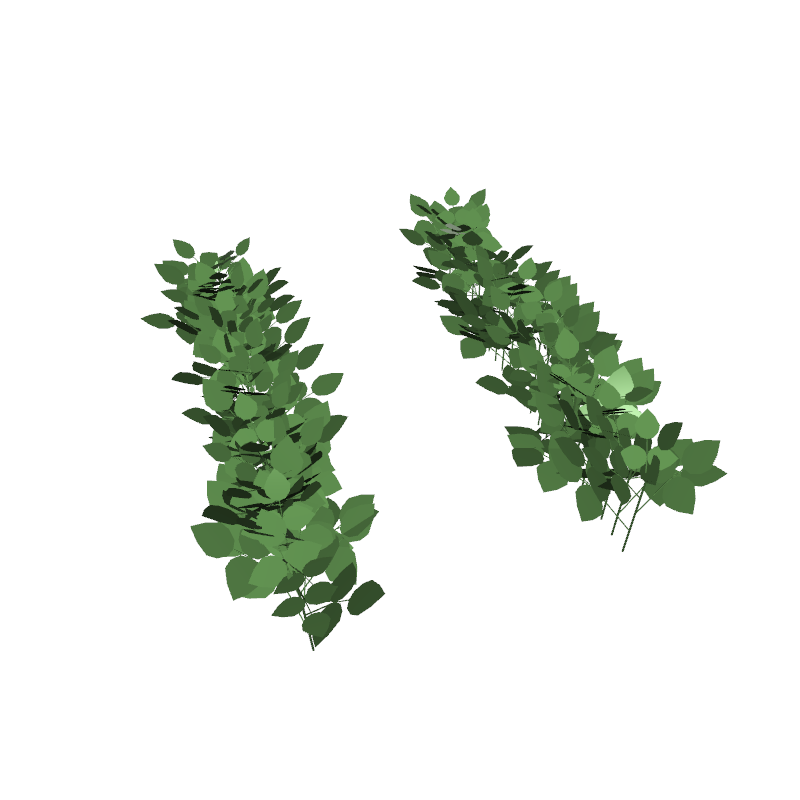}& 
        \includegraphics[width=.19\linewidth]{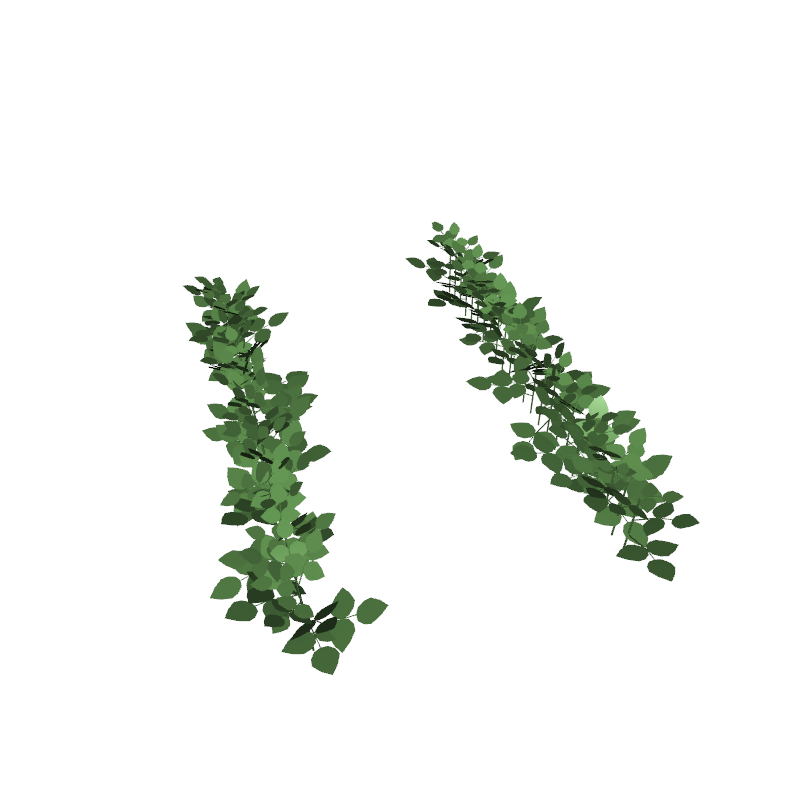}
        \\
          \tikz{
        \node[draw=black, line width=.5mm, inner sep=0pt] 
            {\includegraphics[width=.25\linewidth]{fig/qual_results/input0711.jpg}};
            \node[draw=black, draw opacity=1.0, line width=.3mm, fill opacity=0.8,fill=white, text opacity=1] at (-1.45 , 1.25) { July 11 };
        } &
        \includegraphics[width=.19\linewidth]{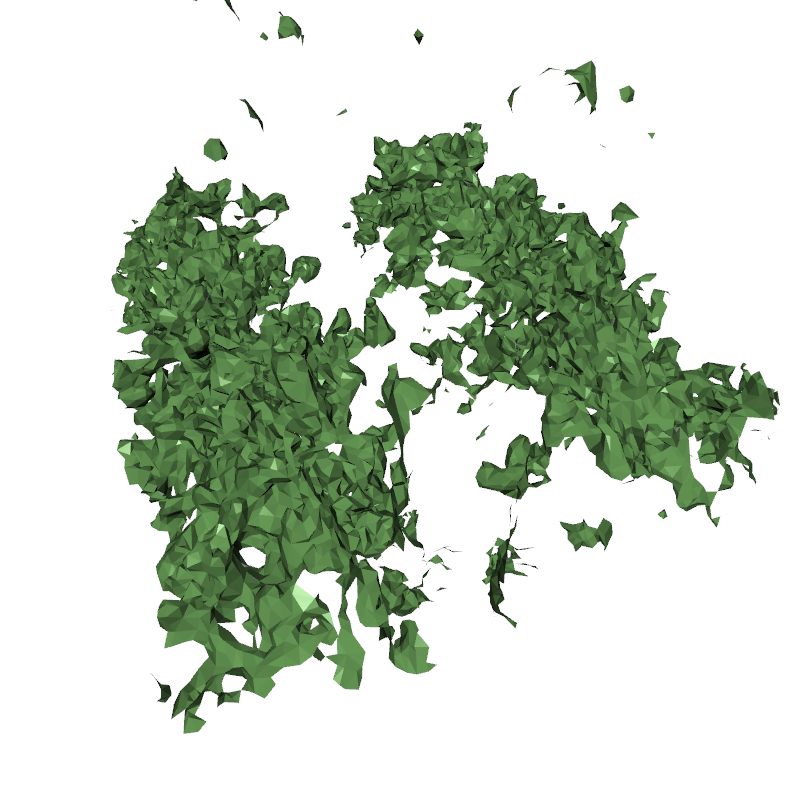}& 
        \includegraphics[width=.19\linewidth]{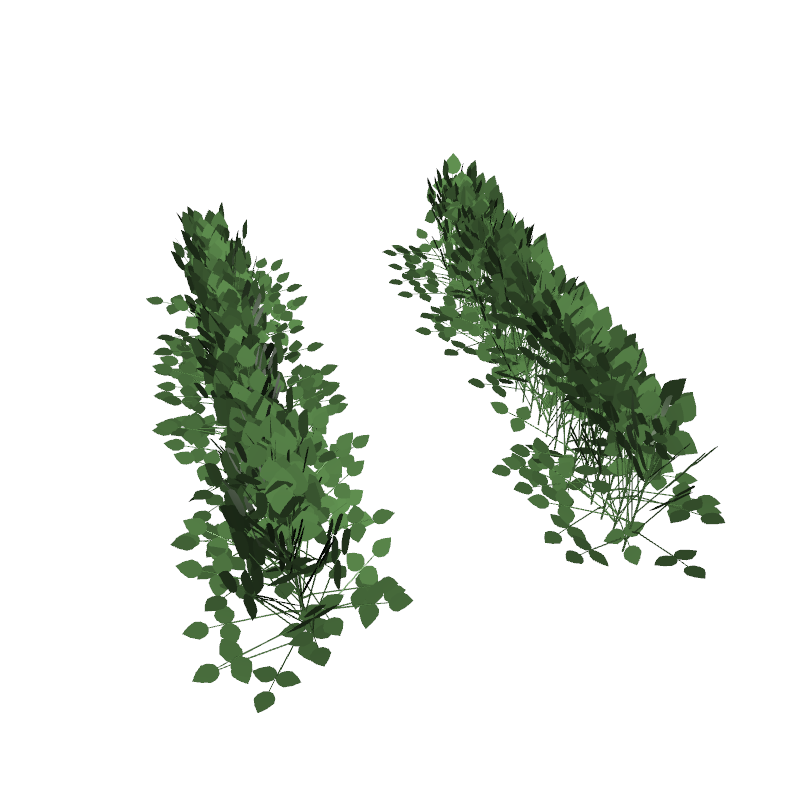}& 
        \includegraphics[width=.19\linewidth]{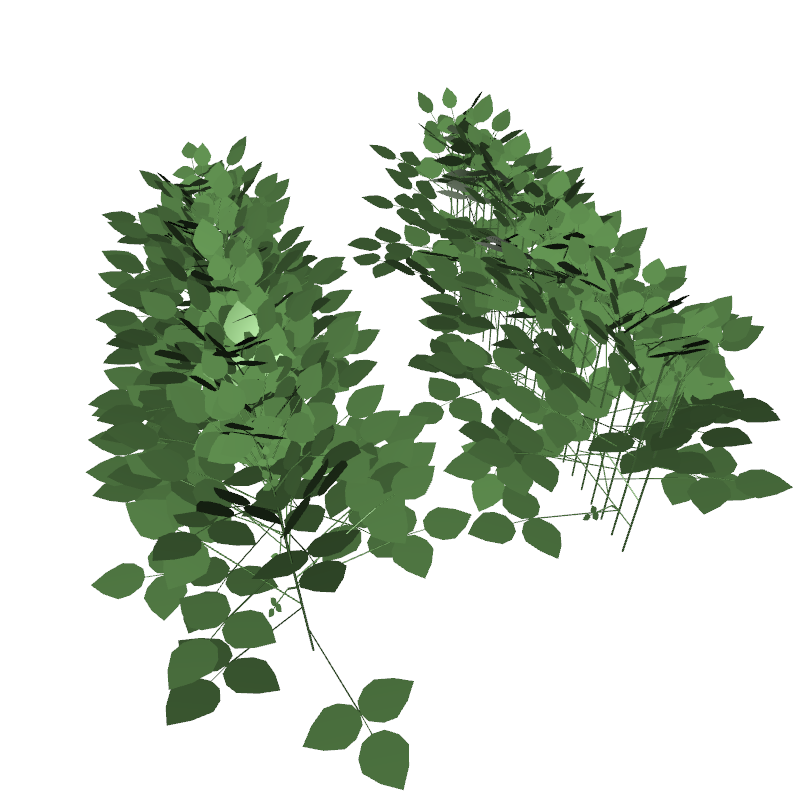}& 
        \includegraphics[width=.19\linewidth]{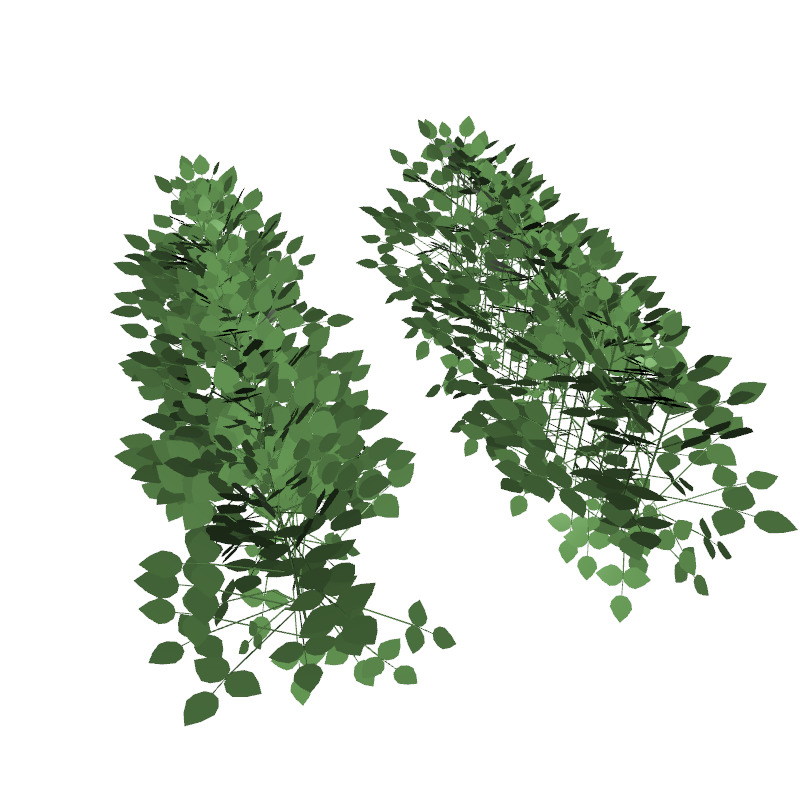}
        \\
    \end{tabular}
    
    }
\captionof{figure}{\textbf{Qualitative reconstruction comparison}. We compare example reconstructions from each of the baseline methods. The first column shows example images from the multi-view data collected at each time point, and the remaining columns show the corresponding mesh reconstructions. 
}
\label{fig:supp_qual_comparison}
\vspace{-10pt}
\end{table*}

\subsection{Additional Row Fitting and Depth Rendering Details}
Our row fitting procedure operates on a point cloud sampled from the NeRF, using the sampling implementation in Nerfstudio~\cite{tancik2023nerfstudio}. For the soybean data, we sample 100K points in a $\SI{2}{\meter} \times \SI{2}{\meter} \times \SI{3}{\meter}$ box centered at $(0, 0, -1.5)$\SI{}{\meter}. Voxel-downsampling is performed at a resolution of \SI{1}{\cm}. Color-thresholding is performed in LAB color space. All points with $L^* < 0$ and $b^* < 1$ are removed. Then, points with $a^* < 2$ are classified as ground points, and the rest are classified as plant points. RANSAC is used to fit a plane to the ground points using an inlier threshold of \SI{5}{\cm} and 1000 max iterations. For row fitting, the points whose distance from the ground are below the 50th percentile are discarded. RANSAC is applied sequentially to fit lines with an inlier threshold of \SI{20}{\cm} until the number of points remaining is either less than 1000 or less than 20\% of the starting number. We then take the row with the most inliers and fit a line through the inliers again using least-squares fitting. The camera pose for depth rendering is set to be directly above the mean of the inlier points at a height of \SI{1}{\meter} from the ground, facing downwards and with the camera's $x$-axis aligned with the row line. If the number of initial plant points is greater than 75\% of the total, we instead use the 70th percentile for distance from the ground, a row inlier threshold of \SI{25}{\cm}, and a render height of \SI{1.25}{\meter}. The render resolution is $994 \times 738$ with a vertical FOV of 50 degrees.

For the maize data, we sample 5M points in a $2 \times 2 \times 2$ box around the origin in the scaled scene. The $L^*, a^*,$ and $b^*$ thresholds are set to 32, 0, and 0 respectively. The ground-inlier threshold is set to \SI{10}{\cm} and the render height is set to \SI{5}{\meter}. The row fitting is done in a ROI with the shape of a cylinder with radius \SI{2}{\meter} whose axis is perpendicular to the ground plane.

\subsection{Mask and Histogram Details}
The foreground mask $\mathbf{M}_{\text{obs}}$ is obtained by color-thresholding on the RGB render combined with thresholding on 3D coordinates.  For soybean, points are marked as background if their $a^*$ channel is less than -8 and their $z$-coordinate is greater than \SI{0.25}{\meter} short of the render height, or their $z$-coordinate is less than \SI{0.1}{\meter}, or their $y$-coordinate is at least \SI{0.5}{\meter} away from zero. For maize, points are marked as background if $a^*$ channel is less than -8 and their $L^*$ channel is greater than 40 and their $z$-coordinate is greater than \SI{2}{\meter} short of the render height, or their $z$-coordinate is less than \SI{0.1}{\meter}, or their $y$-coordinate is at least \SI{3}{\meter} away from zero. $\mathbf{M}_{\text{pred}}$ is obtained trivially by not rendering any background.

For soybean, the depth histogram has 20 equally spaced bins from \SI{0.1}{\meter} to the render height. The lateral histogram has 10 equally spaced bins from zero to half the render height. The depth derivative histogram has 10 equally spaced bins from zero to 0.004. A Gaussian blur with kernel size 25 is applied to the depth map before the calculation of Sobel derivatives, since the NeRF rendering is generally blurrier than the procedural mesh rendering. For maize, the depth histogram has 10 equally spaced bins from \SI{2}{\meter} to the render height, and the blur kernel size is 55.

\subsection{Baseline Implementation Details}
We provide implementation details for the baseline reconstruction methods:

\begin{itemize}
    \item \textbf{Poisson:} We use the Poisson surface reconstruction implementation by Open3D~\cite{zhou2018open3d} integrated into Nerfstudio~\cite{tancik2023nerfstudio}. For soybean, we use 100K points with 50K faces in the same bounding box as in our method and filter out points whose $z$-coordinates are lower than 0.2 plus the 1st percentile. For maize, we use a $\SI{4}{\meter} \times \SI{4}{\meter}$ bounding column and \SI{0.1}{\meter} instead of 0.2.
    \item \textbf{MLP:} The MLP input consists of the same histograms used for optimization, concatenated with the render height and mask area. The training data is generated by uniformly randomly sampling parameter values and computing the corresponding MLP inputs. For soybean, we sample 10K pairs with render height \SI{1.0}{\meter} and 10K pairs with render height \SI{1.25}{\meter}. For maize, we sample 20K pairs with render height \SI{5.0}{\meter}. The MLP has two hidden layers of size 512, and we train for 200 epochs using an Adam optimizer with $\alpha = 0.001, (\beta_1, \beta_2) = (0.9, 0.999)$, and weight decay of $10^{-5}$. 
    \item \textbf{Trust-Region:} We use the trust-region method implemented by the ``trust-constr" option of SciPy's \texttt{optimize.minimize} function~\cite{scipy}, with the maximum number of iterations set to 500. For fair comparison with our Bayesian optimization method, we run it 10 times with different random initializations. We take the solution with the lowest loss value instead of averaging, since we find it gives better performance.
\end{itemize}

\section{Additional Results}
\label{sec:additional results}
\subsection{Additional Qualitative Results}
We provide qualitative comparisons of the different reconstruction baselines in Fig.~\ref{fig:supp_qual_comparison}. We observe that Poisson reconstruction gives incomplete and implausible canopy shape, motivating the use of inverse procedural modeling. Compared to the other procedural generation baselines, our method generally produces the parameters that best reflects the true real-world canopy. In the examples shown, the MLP produces leaves that are too small, and the trust-region optimization produces leaves that are too large.

\begin{figure}[t]
  \centering
  \includegraphics[width=\linewidth]{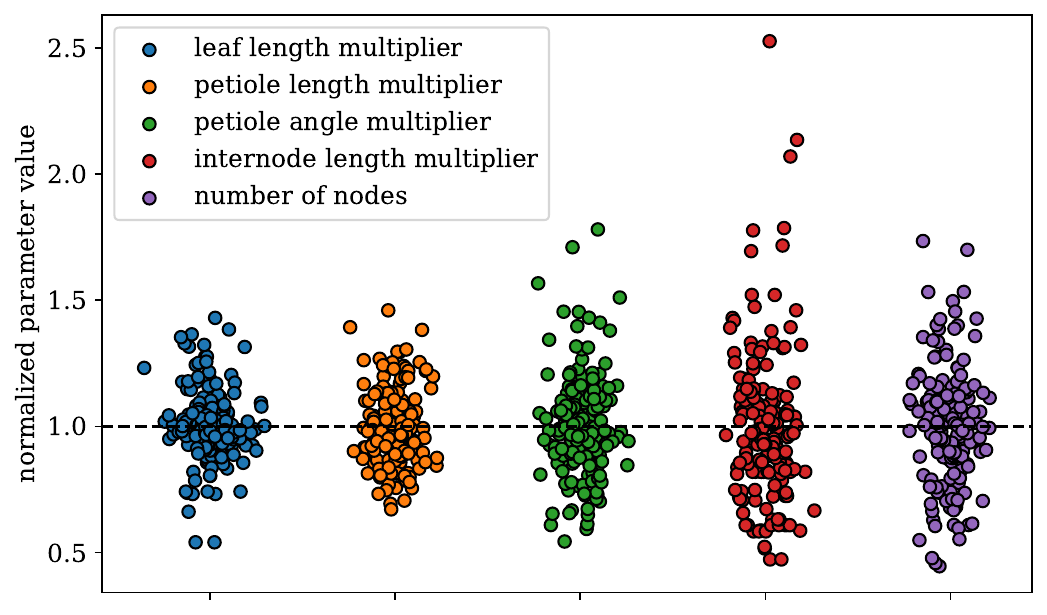}
   \caption{\textbf{Jitter plots of optimized parameter values}. The values are normalized by dividing by the per-scene average. The individual optimization runs generally do not stray far from the average, but taking the average ultimately gives a better result.}
   \label{fig:pa_jitter}
\end{figure}

\subsection{Synthetic Scenes Evaluation}
We perform additional experiments using synthetic scenes to evaluate additional aspects of canopy structure for which it is costly to obtain ground-truth in the real world. We generated 30 synthetic scenes for both soybean and maize by uniformly randomly sampling procedural model parameters and evaluated our method and the Trust-Region baseline on these scenes. We introduce two metrics to measure \textit{stem structure} and \textit{topology}: \textbf{Node Distance Percent Error (NDPE)} is the percent error on the average stem segment length between nodes (branching points), and \textbf{Node Count Percent Error (NCPE)} is the percent error on the total number of nodes. The results (Tab.~\ref{tab:synth}) show that our method can capture these structural variables fairly accurately (but not perfectly).
Note that the Poisson baseline cannot be applied here as it has no stem structure, and the MLP will simply overfit.

\begin{table}[t]
\centering

{
\small
        \begin{tabularx}{0.48\textwidth}{l|>{\centering\arraybackslash}X>{\centering\arraybackslash}X>{\centering\arraybackslash}X|>{\centering\arraybackslash}X>{\centering\arraybackslash}X>{\centering\arraybackslash}X}
        \toprule
        & \multicolumn{3}{c|}{Soybean} & \multicolumn{3}{c}{Maize} \\
        & LAIE $(\downarrow)$& NDPE $(\downarrow)$& NCPE $(\downarrow)$& LAIE $(\downarrow)$& NDPE $(\downarrow)$& NCPE $(\downarrow)$ \\
        \midrule
        
        T-R & 2.49 & 0.48 & 0.91  & 1.46 & 0.13 & 0.70 \\
        Ours & \textbf{0.97} & \textbf{0.21 } & \textbf{0.23}  &   \textbf{0.13 }& \textbf{0.05 }& \textbf{0.08} \\
        \bottomrule
        \end{tabularx}
}
\caption{{\bf Synthetic data evaluation.} {\bf Bold}: best.}
\label{tab:synth}
\end{table}

\subsection{Procedural Model Comparison}
We test the performance of our method with different, simpler procedural models to investigate the importance of the procedural model's design. We implemented the Spherical Crowns and Conical Crowns models from Helios~\cite{bailey2019helios} and fit them using our inverse modeling pipeline. The results (Tab.~\ref{tab:simplified_canopy}) show that while these models can roughly model the visible surfaces, they vastly overestimate the LAI by assuming the space underneath is densely filled with leaves. This confirms the need for grounded morphology models that capture the right assumptions about leaf arrangement.

\begin{table}[t]
\centering

{
\small
        \begin{tabularx}{0.48\textwidth}{l|>{\centering\arraybackslash}X>{\centering\arraybackslash}X|>{\centering\arraybackslash}X>{\centering\arraybackslash}X}
        \toprule
        & \multicolumn{2}{c|}{Soybean} & \multicolumn{2}{c}{Maize} \\
        & LAIE $(\downarrow)$ & LAIPE $(\downarrow)$ &  LAIE $(\downarrow)$ & LAIPE $(\downarrow)$  \\
        \midrule
        
        S-C & 17.85 & 1.42 & 52.12 & 6.74 \\
        C-C & 53.03 & 6.13 & 46.96 & 5.91 \\
        Ours & \textbf{0.69} & \textbf{0.15} & \textbf{0.97} & \textbf{0.26} \\
        \bottomrule
        \end{tabularx}
}
\caption{{\bf Procedural model comparison.} {\bf Bold}: best.}

\label{tab:simplified_canopy}
\end{table}

\subsection{Bayesian Optimization Performance Analysis}
\label{sec:bo_analysis}

We visualize the distributions of the optimized parameter values before averaging in Fig.~\ref{fig:pa_jitter}. The values include the 10 runs for all 18 soybean scenes and are normalized by the averaged result. We observe that there is a moderate amount of variance in the optimization results. We then examine how reconstruction performance changes as we average across different numbers of random initializations for Bayesian optimization in Tab.~\ref{tab:runs_analysis}. Although there is no significant improvement in the leaf angle metrics, we observe that the leaf area estimates improve overall when averaging over more runs, eventually plateauing around 10-20 runs. Note that the constraints of the procedural model generally ensure that all possible combinations of parameters yield plausible 3D reconstructions, so there is no real concern over the average giving an unreasonable result.

\begin{table}[t]
\centering
{
\small

        \begin{tabular}{c|cccc}
        \toprule
         & \multicolumn{4}{c}{Soybean} \\
        \# of runs & LAIE $(\downarrow)$ & LAIPE $(\downarrow)$ & AME $(\downarrow)$ &  ASDE $(\downarrow)$   \\
        \midrule
        1 & 0.76 & 0.23 & 12.18 & 7.81  \\
        2 & 0.79 & 0.20 & 12.18 & 7.74  \\
        5 & 0.78 & 0.16 & 12.13 & 7.74   \\
        10 & 0.69 & 0.15 & \textbf{12.07} & \textbf{7.39}   \\
        20 & \textbf{0.69} & \textbf{0.14 }& 12.23 & 7.77   \\
        \bottomrule
        \end{tabular}
}
\caption{{\bf Effect of averaging solutions over runs.}  {\bf Bold}: best.}
\label{tab:runs_analysis}
\end{table}

We also examine how performance changes as the number of Bayesian optimization iterations varies in Tab.~\ref{tab:iters_analysis}. Here, we are still averaging over 10 runs. In addition to the previous metrics, we report the average value of the minimized loss function and time per run. Performance does tend to improve with more iterations, but takes increasing amounts of time per iteration due to larger and larger kernel computations. In addition, beyond a certain point, decreases in the loss function does not necessarily translate to better reconstruction quality.

\begin{table*}[t]
\centering
{
\small

        \begin{tabular}{c|cccccc}
        \toprule
         & \multicolumn{6}{c}{Soybean} \\
        \# of iterations & LAIE $(\downarrow)$ & LAIPE $(\downarrow)$ & AME $(\downarrow)$ &  ASDE $(\downarrow)$ & Average $L (\downarrow)$ & Time $
        (\downarrow)$ \\
        \midrule
        300 & 0.73 & 0.18 & 12.32 & 7.77 & 0.0172 & \textbf{3m} \\
        400 & 0.77 & 0.17 & 12.18 & 7.77 & 0.0133 & 9m \\
        500 & \textbf{0.69} & 0.15 & \textbf{12.07} & \textbf{7.39} & 0.0110 & 16m  \\
        600 & 0.70 & \textbf{0.13} & 12.56 & 8.01 & \textbf{0.0096} & 26m  \\
        \bottomrule
        \end{tabular}
}
\caption{{\bf Effect of number of Bayesian optimization iterations.}  {\bf Bold}: best. }
\label{tab:iters_analysis}
\end{table*}

\begin{figure*}[!t]
\centering
\includegraphics[width=\linewidth, trim={0 14cm 0 0cm}, clip]{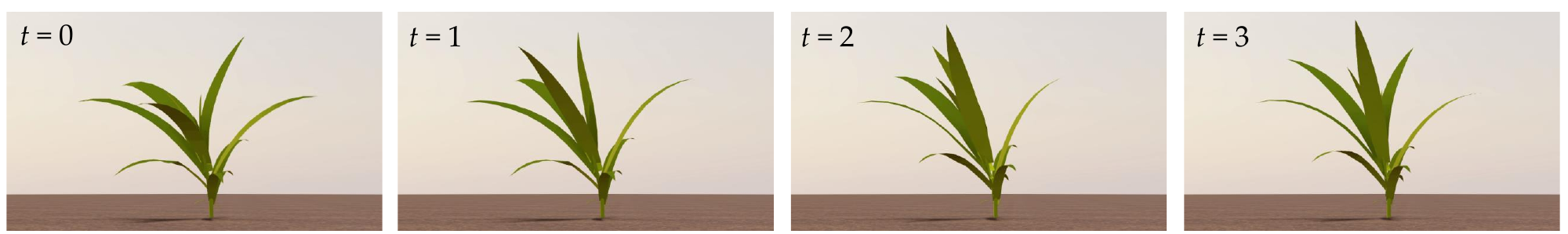}
\vspace{-5mm}
\caption{\textbf{Wind simulation.} We show that the procedurally generated mesh outputs can be used for physically realistic dynamics simulations. Here, a uniform wind force is applied directly from the right. Video visualizations are included in the supplementary video.}
\vspace{-2mm}
\label{fig:wind}
\end{figure*}

\subsection{Wind Simulation}
The procedurally generated meshes from our model can also be used for dynamics simulations. We show rendered frames of a wind simulation using NVIDIA PhysX in Fig.~\ref{fig:wind}. Visually, the results appear to show a physically plausible animation of leaves blowing in the wind. Video visualizations can be found in the supplementary video.

\section{Limitations}
\label{sec:limitations}
One limitation of our method is that the inverse procedural modeling is dependent on the NeRF reconstruction performance, which may degrade due to wind-induced leaf motion. Another limitation is that the procedural generation models used cannot model certain details, e.g. damaged leaves and non-leaf organs of the plants. A final limitation is that our RANSAC-based row-fitting method is sensitive to hyperparameters such as the inlier threshold and segmentation threshold, which may need to be tuned per dataset. Improving the robustness by leveraging learning-based methods is a potential direction for future work.

\end{document}